\documentclass{article}

\usepackage{arxiv}

\usepackage[utf8]{inputenc} 
\usepackage[T1]{fontenc}    
\usepackage{hyperref}       
\usepackage{url}            
\usepackage{booktabs}       
\usepackage{amsfonts}       
\usepackage{nicefrac}       
\usepackage{microtype}      
\usepackage{lipsum}
\usepackage{authblk}

\usepackage{wrapfig}
\usepackage{balance}
\usepackage[numbers]{natbib}
\usepackage{float} 
\usepackage{times}
\usepackage{epsfig}
\usepackage{amsmath, amsfonts, amssymb}
\usepackage{bm}
\usepackage{algorithm}
\usepackage{algorithmicx}
\usepackage{comment}
\usepackage{enumitem}
\usepackage{fullpage}
\usepackage{tikz}
\usepackage{pifont}
\usepackage{amsthm}
\usepackage{subcaption}
\usepackage{mathtools}
\usepackage{multirow}
\usepackage[noend]{algpseudocode}
\usepackage{booktabs}
\usepackage[export]{adjustbox}
\usepackage{url}

\usepackage{xcolor}
\hypersetup{
    colorlinks=true,
    linkcolor=blue,
    citecolor=blue,
    urlcolor=blue
}

\usepackage[textsize=scriptsize,backgroundcolor=green]{todonotes}

\newtheorem{theorem}{Theorem}

\newtheorem*{remark}{Remark}

\newcommand{\cmark}{{\color{blue}\ding{51}}}%
\newcommand{\xmark}{{\color{red}\ding{55}}}

\newcommand\numberthis{\addtocounter{equation}{1}\tag{\theequation}}

\newcommand\Mycomb[2][^n]{\prescript{#1\mkern-0.5mu}{}C_{#2}}

\usepackage{xspace}
\newcommand*{\eg}{\textit{e.g.},\@\xspace}
\newcommand*{\ie}{\textit{i.e.},\@\xspace}
\newcommand*{\sut}{\textit{s.t.}\@\xspace}
\newcommand*{\vs}{\textit{vs.}\@\xspace}
\newcommand*{\wrt}{\textit{w.r.t.}\@\xspace}
\makeatletter
\newcommand*{\etc}{%
	\@ifnextchar{.}%
	{\textit{etc}}%
	{\textit{etc.}\@\xspace}%
}
\makeatother
\algtext*{EndWhile}
\algtext*{EndFor}
\algtext*{EndIf}
\algdef{SE}[DOWHILE]{Do}{doWhile}{\algorithmicdo}[1]{\algorithmicwhile\ #1}%

\makeatletter
\def\BState{\State\hskip-\ALG@thistlm}
\makeatother

\title{Fairness-Aware Multi-Group Target Detection\\ in Online Discussion}

\author[1]{\textbf{Soumyajit Gupta}}
\author[2]{\textbf{Maria De-Arteaga}}
\author[3]{\textbf{Matthew Lease}}
\affil[1]{Dept.\ of Computer Science, The University of Texas at Austin}
\affil[2]{Department of Data, Analytics, Technology, and Artificial Intelligence, ESADE}
\affil[3]{The Information School, The University of Texas at Austin}
\affil[ ]{\texttt{\small 
\{$^1$smjtgupta,$^3$ml\}@utexas.edu, $^2$maria.dearteaga@esade.edu}}

\begin{document}

\maketitle

\begin{abstract}

Target-group detection is the task of detecting which group(s) a piece of content is ``directed at or about''. Applications include targeted marketing, content recommendation, and group-specific content assessment. Key challenges include: 1) that a single post may target multiple groups; and 2) ensuring consistent detection accuracy across groups for fairness. In this work, we investigate fairness implications of target-group detection in the context of toxicity detection, where the perceived harm of a social media post often depends on which group(s) it targets. Because toxicity is highly contextual, language that appears benign in general can be harmful when targeting specific demographic groups. We show our {\em fairness-aware multi-group target detection} approach both reduces bias across groups and shows strong predictive performance, surpassing existing fairness-aware baselines. To enable reproducibility and spur future work, we share our code online at \href{https://github.com/smjtgupta/Multi-GAP}{https://github.com/smjtgupta/Multi-GAP}. \footnote{Paper accepted as part of The 2026 ACM Conference on Fairness, Accountability, and Transparency (FAccT '26)}
\\

\end{abstract}

\section{Introduction}
\label{sec:intro}

In online discussions, properly interpreting the discourse and managing content requires knowing which groups(s) are being referred to in discussion. {\em Target-group detection} is the task of detecting which group(s) a piece of content is ``directed at or about'' \cite{sachdeva2022assessing}. Applications include targeted marketing, content recommendation, and group-specific content assessment. In the first two cases, the group that may deem certain content relevant is the target. For content assessment, the group that a post speaks to or about is the target.

In fact-checking, identifying a post's target is key to assessing the post's \emph{checkworthiness} (\ie appropriately prioritizing it for fact-checking) since ``potential harm to specific groups" can be one of the criteria used for prioritization~\citep{neumann22,liu2024exploring}. 
In toxicity detection, whether a post is hateful or discriminatory is often highly contextual, with language that might seem benign taking on harmful undertones if it targets a specific group \cite{davidson2019racial,sap2019risk,markl2022language,kennedy2020contextualizing,breitfeller2019finding}. 
Target-group detection is thus \textbf{important}, \textbf{distinct}, and \textbf{separable} from any downstream tasks based on detected targets, such as  assessing if a post is toxic~\citep{gupta2023same} depending on the target(s). Beyond moderation, fair and accurate target-group prediction also has other use cases. For example, social science studies examining the impact of harmful language toward different communities rely on correct group attribution \citep{argyle2024misclassification,lockhart2023name,imana2025auditing}, and disparities in such detection would lead to distorted empirical observations.


This problem poses two key challenges.
\underline{First}, \textit{a single post may target multiple groups}. If we consider toxicity detection, prior work has largely framed target-group detection as a single-label task, assuming each post targets at most one group \cite{blodgett2017dataset,sap2019risk}. Most toxicity datasets \cite{waseem2016hateful,davidson2017automated} also lack granular demographic annotations or focus narrowly on binary classifications. This fails to account for the complex interplay of cultural, linguistic, and identity-based differences in what constitutes harmful content. It also misses the multi-label nature of real-world discourse, where language can implicate multiple groups simultaneously, as illustrated below in \textbf{Table \ref{tab:task-setup}}. 
\underline{Second}, \textit{it is not enough to detect groups accurately; we must also detect groups fairly} for equitable treatment across diverse demographics. For example, in personalized content recommendation, such as suggesting movies or music to diverse user groups, accuracy disparities in target group detection could result in one or more groups failing to receive content that is relevant to them. In toxicity detection, disparities in target group detection can have downstream detrimental impacts on toxicity detection models that assess toxicity conditional on the target group \citep{gupta2023same,fleisig2023majority}.

\begin{table*}[ht]
    \centering
    \vspace{-0.5em}
    \caption{Sample comments from the MHS Corpus \cite{sachdeva2022assessing}, showing the multi-label format for groups. Note that groups can be targeted by a post regardless of the post's  toxicity. We have partially redacted toxic language in examples shown.}
    \vspace{-0.75em}
    \resizebox{\textwidth}{!}{
    \begin{tabular}{p{10.9cm}|c|ccccccc} \toprule
        \textbf{Post} & \textbf{Toxic} & \multicolumn{7}{c}{\textbf{Demographic Groups Targeted \textnormal{(\ie group(s) the post is ``directed at or about'')}}} \\ 
        & & Asian & Black & Latinx & Middle Eastern & Native American & Pacific Islander & White \\ \midrule
        I'm chinese but have really dark skin due to my parent's genes, and I agree, my country's (singapore) people are all friends with different races and religions. & \xmark & \cmark & \xmark & \xmark & \xmark & \xmark & \xmark & \xmark\\ \midrule
        There is plenty of evidence that women and non-white people are disadvantaged in almost every hiring process in the western world. & \xmark & \cmark & \cmark & \cmark & \cmark & \cmark & \cmark & \xmark\\ \midrule
        ... go back to the sh** hole you crawled out of. You America hating jihadi. & \cmark & \xmark & \xmark & \xmark & \cmark & \xmark & \xmark & \xmark \\ \bottomrule
    \end{tabular}
    }
    \label{tab:task-setup}
    \vspace{-0.75em}
\end{table*}

\textbf{The value of symmetric error weighting.} To date, most fairness literature has focused on 
scenarios for allocating goods or burdens (\eg college admissions or bail decisions) wherein false positives and negatives have asymmetric costs \citep{kleinberg2016inherent}. \textit{Target-group detection is different.} Consider content recommendation, where preferences (\eg in movies or music) vary across groups. If our goal is to promote fairness across all groups, there is no universal notion of ``positive" and ``negative" class because content relevant to one group may be irrelevant to another, and all groups should be equally likely to encounter relevant content. Symmetric error weighting across groups treats groups equitably.

\textit{The same principle holds in toxicity detection}, which can also be viewed as a task of allocating goods and burdens. From the perspective of targets, content moderation that protects the target from toxic content is a good, and so gaps in false negatives in toxicity detection may constitute an issue of distributive justice~\citep{neumann22,gupta2023same}. In contrast, the upstream task of target-group detection is a multi-label task: labels are demographic groups themselves (\eg Latinx, Black), and errors misidentifying one group for the other are equally undesirable. For instance, mistaking a post targeting Latinx as one targeting Black is equally undesirable as the reverse. 

\textbf{How can we optimize for symmetric errors across groups?} The Accuracy Parity (AP) measure \cite{zhao2020conditional} provides a natural fairness criterion for the task of multi-target group detection. To optimize for AP during model training, we extend the Group Accuracy Parity (GAP) loss function \cite{kovatchev2022fairly}, which penalizes accuracy deviations across groups. Specifically, because GAP is limited to binary settings, \textbf{we introduce} {\boldmath $GAP_{multi}$}, a scalable loss that extends GAP to enable fair detection when posts target multiple groups. Our formulation enables fair optimization in realistic multi-group settings while preserving the convergence behavior of the original GAP loss.

\textbf{Why not use Equalized Odds (EO)?} While EO might be considered instead of  AP to independently enforce equality in both error types, we show that EO \cite{hardt2016equality} {\em does not actually treat both types of errors equally}, and as a consequence, gives priority to some groups over others. Through a formal theoretical analysis, we present an \textit{impossibility result}: under realistic scenarios (\eg unequal group base rates), one cannot simultaneously satisfy both AP and EO. Moreover, we empirically show that enforcing EO can actually degrade performance for some groups. Consequently, its use in target detection setting could lead to unintended consequences. 

\textbf{Evaluation context: toxicity detection}. Having motivated \textit{fairness-aware multi-group target detection} across several domains, we now focus our empirical study in the context of toxicity detection. Note that \textit{we  \underline{do not} evaluate toxicity itself in this study}. We only evaluate the preceding task of multi-group target detection. As discussed earlier (paragraph 2), identifying the target of a post is a necessary precursor to assessing if a post is toxic or not. 

We use the large-scale \textit{Measuring Hate Speech} (MHS) Corpus \cite{sachdeva2022assessing} and \textit{HateXplain} \cite{mathew2021hatexplain} datasets for our study because these: 1) include and annotate posts targeting multiple groups; 2) annotate targets for \underline{both} toxic and non-toxic posts (a post can be ``directed at or about'' (Ibid.) one or more target group(s) without being toxic); and 3) include demographic group annotations: race, gender, religion, among others.  Evaluation on these two datasets, which include data from multiple platforms (Twitter, Reddit, YouTube, Gab), validates generalization across different discourse styles and moderation environments, testing robustness beyond platform-specific artifacts.

We compare our proposed multi-group $GAP_{multi}$ loss to several baseline loss functions, including traditional fairness-aware losses and standard optimization strategies. \textbf{We assess two key objectives}: \textbf{group-fairness} (balanced performance across demographic groups) \textbf{and utility} (overall predictive accuracy). $GAP_{multi}$ not only achieves better group-fairness (by minimizing performance disparities impact across groups), but also maintains competitive overall accuracy. Our work thus effectively addresses the unique challenges of {\em multi-group target detection}, consistently reducing bias across groups while maintaining similar utility.

\textbf{Summary of contributions}: We motivate and propose a \textit{fairness-aware multi-group target detection} framework that jointly optimizes overall utility and group fairness in a multi-label setting. The proposed $GAP_{multi}$ loss enables parallel GPU computations of group-pairwise errors, allowing a constant factor scaling to an arbitrary number of groups. This supports scalable optimization needed for deployment in dynamic online platforms. Empirical validation across datasets demonstrate the robustness of our loss function, bridging theoretical fairness objectives with real-world applicability. This provides a robust and practical solution for group-identification tasks by making predictions both fairly and accurately. To spur future research and support  benchmarking, we also share our code online (see Abstract).

\section{Related Work} \label{sec:litrev}

\textbf{Fairness in Machine Learning}
has been studied across critical applications such as risk assessment \cite{hardt2016equality}, college admissions \cite{kleinberg2016inherent}, loan approval \cite{chouldechova2017fair}, and recidivism \cite{angwin2022machine}. These domains typically involve tasks where asymmetric errors, \ie allocating goods or burdens, are central to the decision-making process. For instance, in loan approvals, false negatives (denying credit to a qualified individual) may carry a greater societal cost than false positives (approving a loan for someone who defaults). Similarly, in recidivism prediction, false positives (predicting someone will re-offend when they would not) disproportionately impacts marginalized groups, making error prioritization critical. For this reason, fairness measures such as Equalized Odds~\cite{hardt2016equality} and Demographic Parity \cite{zemel2013learning} are often employed. Given their focus on error rate alignment across groups, these measures are well-suited for tasks where the interpretation of error rates is tied to the societal implications of unequal access to resources or unjust burdens. 

In contrast, \textbf{multi-group target detection} offers a different fairness challenge wherein errors are symmetric: misidentifying a targeted group (false negative) and incorrectly assigning a group (false positive) are equally undesirable. Such symmetry is relatively rare in classic fairness applications but critical here, motivating our focus on Accuracy Parity (AP) \cite{zhao2020conditional} as a natural and principled measure. While AP has been studied conceptually \cite{berk2021fairness,das2021fairness,mitchell2021algorithmic}, relatively little work has explored it for multi-group detection. 

In toxicity modeling, identifying which demographic group(s) are targeted by toxic language has become a crucial component in modern toxicity detection pipelines. Early work by \citet{blodgett2017dataset} framed target-group detection as a single-label classification task across four groups, selecting the most likely group during inference. Similarly, many datasets  \cite{davidson2017automated,fortuna2018survey} treat group identification as binary or mutually exclusive, assuming each post targets only one group. However, real-world toxic content often implicates multiple communities simultaneously. While recent studies discuss the need the considering group-targeted toxicity \cite{vidgen2020directions} or seek to model it \cite{kumar2021designing,gordon2022jury}, to the best of our knowledge, all have assumed oracle detection of target groups based on annotated corpora.


A central challenge in fairness-aware machine learning is aligning the training objective (\ie the loss function) with the fairness criterion used for evaluation. When this alignment is weak, as in the case of many surrogate loss functions \cite{xia2020demoting,yu2024fairbalance,chen2024hate}, models may experience \textit{metric divergence}, especially under sensitive hyperparameter regimes \cite{morgan2004direct}. 
These surrogate losses typically optimize for proxies (\eg cross-entropy with reweighting) rather than directly minimizing fairness violations. In contrast, \textit{strictly-mapped loss functions} \cite{kovatchev2022fairly,shen2022optimising} are constructed to directly correspond to fairness measures. These formulations offer a smoother and more reliable optimization path --- as the loss decreases, the corresponding fairness metric improves predictably.

Group-specific toxicity often manifests through distinct language patterns: coded language targeting certain ethnicities or innocuous terms as racialized slurs in  context. For example, ``wet back" may be used in a factual description, but it can also be used as a slur against Latinx people. Gender-based hostility often involves stereotype-driven insults, where superficially benign content, may acquire harmful undertones when directed at particular groups. Without detecting the intended target groups, toxicity detection models risk generalization, causing misclassifications and disparate impact for minority groups \cite{sap2019risk,davidson2019racial}. Target labels from annotators enable group-conditioned toxicity detection models \cite{gordon2022jury,fleisig2023majority,gupta2023same}, that capture nuances between shared and group-specific toxicity patterns. However, real-world deployment requires detecting target groups in-the-wild, as content lacks pre-annotated labels. While user reports may provide this data, proactive detection demands an accurate and fair target-detection module in place.


\section{Multi-Group Target Detection} \label{sec:probstat}

\textbf{Problem Statement.} 
Given a collection of $N$ posts, we represent each post as a feature vector in $\mathbb{R}^F$ using a pre-trained text encoder, yielding a dataset $X \in \mathbb{R}^{N \times F}$. 
For a set of demographic groups $G$ (\eg race, religion, gender), let the label space be $Y \in \mathbb{R}^{N \times |G|}$. Each post $x_i$ has a corresponding multi-hot label vector $y_i=\{0,1\}^{|G|}$, where $y_i[g]=1$
indicates that group $g \in G$ is targeted ($1$) or not ($0$) by post $x_i$. This yields a multi-label classification problem \cite{mccallum1999multi, herrera2016multilabel} where a single post $x_i$ can target multiple groups. We seek to learn a prediction function $f: X \rightarrow \{0,1\}^{|G|}$ mapping each post $x_i$ to its multi-group associated label $y_i$. 

We report group-specific error $err_g$ for each group $g$, and overall (macro-average) error over all groups, $\overline{err}_{overall}$.
%
From a utility perspective, we aim to minimize this $\overline{err}_{overall}$. 
However, to ensure fairness, we also seek to minimize disparities in predictive performance across groups. We define the inter-group disparity ($dev_{overall}$) as the summation of deviation of each group's error from the mean overall error for predicted labels $\hat{Y}$. Thus, we have the following set of optimization problems in \textbf{Eq.~\ref{eq:overall}} and \textbf{Eq.~\ref{eq:deviation}}: 

\begin{equation}
    \underset{\hat{Y}}{\min}\,\,\overline{err}_{overall}
    = \frac{\sum_{g=1}^{|G|} err_g}{|G|}
    \label{eq:overall}
\end{equation}
\begin{equation}
    \underset{\hat{Y}}{\min}\,\, dev_{overall}
    = \sum_{g=1}^{|G|} |err_g - \overline{err}_{overall}|
    \label{eq:deviation}
\end{equation}


To jointly optimize for both utility and group fairness, we adopt a Single Objective Optimization (SOO) goal, combining these objectives, Eq.~\ref{eq:overall} as primary and Eq.~\ref{eq:deviation} as secondary, with a regularization term $\lambda \in (0, \infty)$ in \textbf{Eq.~\ref{eq:soo}}. This formulation explicitly balances model utility with group-level fairness, penalizing uneven performance across groups.
\begin{align}
    \text{\textbf{SOO goal:}} \qquad \underset{\hat{Y}}{\min} \,\, (\overline{err}_{overall} + \lambda \cdot dev_{overall} )\label{eq:soo}
\end{align}

\section{From Measure to Optimizer: AP \& GAP}

Achieving fairness in target-group detection requires not only principled metrics to measure fairness but also optimization strategies that enforce these fairness objectives. 
In this section, we discuss both Accuracy Parity (AP) \cite{zhao2020conditional} as a fairness evaluation measure for target-group detection, as well as Group Accuracy Parity (GAP) \cite{kovatchev2022fairly} as its optimization-aligned counterpart.

\textit{Accuracy Parity (AP).} To balance 
accuracy \vs group fairness, AP measures the degree to which predictive accuracy varies across demographic groups. 
%
As discussed in Section~\ref{sec:intro}, AP is well-aligned with multi-group target detection because it \textit{aligns with the symmetric error costs inherent in the task}, treating errors from all groups equally, irrespective of group membership (Eq.~\ref{eq:deviation}). 
For instance, misclassifying a post targeting one group as targeting another group is equally undesirable, regardless of the directionality of the error. A particular attribute of the target-group detection task also merits note: \textit{the group and the label are the same}. 


\textit{Group Accuracy Parity (GAP).} To directly optimize for Accuracy Parity (AP), we adopt  \textit{Group Accuracy Parity} (GAP) loss \cite{kovatchev2022fairly}, the only differentiable loss function for AP that we are aware of. Unlike surrogate losses that loosely correlate with fairness metrics, GAP strives to minimize the loss directly to improve AP during training. The GAP formulation shown in \textbf{Eq.~\ref{eq:gap}} combines the overall error (OE) with a regularization term measuring the disparities in cross entropy (CE) error across binary groups ($g=0,1$). It allows different weighted variants of accuracy \wrt chosen entropy, depending on the evaluation need.
\begin{align}
    GAP = OE + \lambda \| \underbrace{CE(g=1)}_\text{err Group 1} - \underbrace{CE(g=0)}_\text{err Group 0} \|_2^2
    \label{eq:gap}
\end{align}
The regularization term evaluates to zero when errors across the two groups are same, \ie perfect AP score of 0. The squared two-norm formulation supports a smooth loss surface resulting in better optimization trajectory. While formulated for binary group setting, this structure forms the foundation for our multi-label extension, as discussed next.

\section{GAP for Multiple Targets - $GAP_{multi}$}

Given the multi-label nature of multi-group target detection (\ie a single post can target 0-all groups), we propose a mathematical extension of the GAP loss named $GAP_{multi}$. 
While we could have followed the Single Objective Optimization (SOO) format in Eq.~\ref{eq:soo}, we seek to address several issues associated with it. We can see that the Overall loss in Eq.~\ref{eq:overall} can be computed in parallel over all the $G$ groups. 
However, Eq.~\ref{eq:deviation} acts as the serial step while computing the mean error value, only after which
the group-level deviations be measured. This deviation-from-mean computational approach provides a serial bottleneck in the otherwise parallel pipeline. To bypass this serial block and improve parallelization, we formulate our $GAP_{multi}$ in the form in \textbf{Eq.~\ref{eq:multigap}}.
\begin{align}
	GAP_{multi} = OE + \lambda \underset{j,k \in G ,j \neq k}{\sum}\|  \underbrace{CE(g=j)}_\text{err Group j} - \underbrace{CE(g=k)}_\text{err Group k}  \|_2^2
    \label{eq:multigap}
\end{align}
The regularization term in Eq.~\ref{eq:multigap} considers pairwise error differences 
for all distinct $\Mycomb[|G|]{2}$ group pairs $j,k \in G ,j \neq k$. This allows for direct comparisons between groups, ensuring that no group is unduly favored or marginalized relative to another. In contrast, deviation-from-mean approaches measure errors against a global aggregate (Eq.~\ref{eq:deviation}), which can mask disparities between individual pairs. For instance, significant differences in performance between the statistical minority groups may remain undetected if their combined performance aligns with the statistical majority group’s average error. The pairwise structure enforces symmetric treatment of errors across all groups. 
The algorithmic flow is described in details via \textbf{Alg. \ref{alg:ovloss}} and \textbf{Alg. \ref{alg:gaploss}} below.

\vspace{-0.5em}

\begin{algorithm}[h]
\small
	\caption{$overall\_loss(y\_true, y\_pred, w[g])$\\ Loss function for optimizing Overall Error}
	\begin{algorithmic}[1]
	\BState \textbf{Input}: $y\_true, y\_pred$ \Comment{True and Predicted Labels}
    \BState \textbf{Input}: $w[g]$ \Comment{Balanced weights of Group $g$}
    \State $y\_true\_lab = y\_true[:\,,0]$ \Comment{Label Info}
    \State $y\_true\_dem = y\_true[:\,,1]$ \Comment{Demographic Info}
    \For {each group $grp[g] \in G$}
        \State $pos\_grp[g] = group(y\_true\_dem == g)$ \Comment{Find indices}
        \State $y\_true[g] = y\_true\_lab[pos\_grp[g]]$ \Comment{True group labels}
        \State $y\_pred[g] = y\_pred[pos\_grp[g]]$ \Comment{Predicted group labels}
        \State $err\_grp[g] = bce(y\_true[g], y\_pred[g], w[g])$ \Comment{wBCE}
    \EndFor
    \State $err\_overall = \sum^{|G|} err\_grp[g]$ \Comment{summation of group loss}
	\BState \textbf{Output}: $err\_overall$ 
	\end{algorithmic} \label{alg:ovloss}
\end{algorithm}

\vspace{-0.5em}

Finally, our $GAP_{multi}$ loss retains the overall error (OE) term (as in Eq.~\ref{eq:soo}) to preserve utility, but it augments OE with a pairwise regularization component enabling parallelizable fairness enforcement across $\Mycomb[|G|]2$ or $|G|(|G|-1)/2$ group pairs. These pairwise error computations, despite quadratic growth with the group cardinality $|G|$, are independent and can be performed in parallel. The formulation is thus computationally scalable, even for large datasets with many demographic groups. Thus it can be integrated seamlessly into modern deep learning pipelines for efficient optimization.

\textbf{Alg. \ref{alg:ovloss}} presents the logic of the overall weighted Binary Cross Entropy (WBCE) loss, where the weights $w[g]$ represent inversely scaled values of labels (0s and 1s), within each group $g$. Thus $err\_grp[g]$ equivalently maps to Balanced Accuracy (BA), an evaluation measure widely used in datasets with label imbalance. Absence of this weighting term would strictly map to standard accuracy, following the mapping from $BCE$. The overall error $err\_overall$ is then defined as summation of errors across groups. Note that we do not weight groups while adding their errors since: a) we want to treat all groups equally; and b) TensorFlow's $bce$ function is scale independent, \ie it produces same error value for equal ratios of mispredictions \wrt total samples, irrespective of sample size. For \eg $bce$ value over 5 samples with 1 misprediction is equal to $bce$ value over 15 samples with 3 misprediction.

\vspace{-0.5em}

\begin{algorithm}[htb]
\small
	\caption{$gap\_loss(y\_true, y\_pred, w[g])$\\ Loss function for optimizing Group Accuracy Parity Error}
	\begin{algorithmic}[1]
	\setcounter{ALG@line}{-1}
    \State \textbf{Repeat Steps 1-10} of \textbf{Alg. \ref{alg:ovloss}}
    \For {group pairs $[j,k] \in G, [j \neq k]$} \Comment{$\Mycomb[|G|]2$ iterations}
        \State $err\_group\_pairs = \sum (err\_grp[j] - err\_grp[k])^2$
    \EndFor
    \State $err\_balanced = err\_overall + \lambda \cdot err\_group\_pairs$
	\BState \textbf{Output}: $err\_balanced$ 
	\end{algorithmic} \label{alg:gaploss}
\end{algorithm}

\vspace{-0.5em}

\textbf{Alg. \ref{alg:gaploss}} presents the logic of our proposed $GAP_{multi}$ loss. After computing overall error via Alg. \ref{alg:ovloss}, $GAP_{multi}$ computes the $\Mycomb[G]{2}$ group-pairwise errors. The errors are squared to enforce positive values and allow for a smooth loss surface. The final SOO error as per Eq. \ref{eq:soo} is the summation of the overall error and a regularized sum of group-pair errors.

\section{Impossibility Results} \label{sec:impossible}

Achieving fairness in target-group detection requires accounting for both false positives (type-I errors) and false negatives (type-II errors) because both can represent harm to marginalized groups. One might consider Equalized Odds (EO) \cite{hardt2016equality} or its surrogates \cite{yu2024fairbalance,chen2024hate} as a viable optimizer. However, as noted in Section~\ref{sec:litrev}, EO is not well-suited for this task because it induces asymmetric errors when base rates differ across groups. We now present a theoretical impossibility result demonstrating that EO and Accuracy Parity (AP) cannot generally be satisfied simultaneously unless strong assumptions (\eg equal base rates) hold. This challenges the common misconception that EO implicitly ensures parity in group-level predictive accuracy. 

The theorems are stated below, with their corresponding proofs in \textbf{Appendix \ref{app:proof}}. Furthermore, we highlight that in datasets with base rate disparities, enforcing EO might severely reduce the predictive accuracy for statistical minorities, at the cost of balancing error rates. For target-group detection, a \textbf{positive} / \textbf{negative} label for each group indicates that group being targeted / not targeted. Thus a false positive (or false alarm) indicates incorrectly detecting a group as being targeted, while a false negative (or miss) indicates failing to detect a group as being targeted.





\begin{theorem}
    A fairness problem aiming to simultaneously satisfy Equalized Odds and Accuracy Parity is only feasible when: 1. the base rates are equal across all groups; or 2. the model engages in random prediction.
\end{theorem}

This theorem highlights the limitations in attempting to satisfy both EO and AP simultaneously. When base rates differ across groups, a common scenario in real-world datasets, achieving parity in error rates (EO) inherently conflicts with maintaining balanced predictive accuracy (AP). The only scenarios where these objectives can coexist are either under the idealized condition of equal base rates ($\delta^*_A = \delta^*_B$), which rarely holds in practice, or when the model resorts to random predictions, which sacrifices utility altogether. 


\begin{theorem}
    A fairness problem aiming to simultaneously satisfy False Positive and Negative Error Disparity (FPNED) and Accuracy Parity is only feasible when the base rates are equal across groups.
\end{theorem}

This theorem states that the alternative fairness measure FPNED \cite{chen2024hate} is incompatible with AP as well for most real-world datasets when base rates are unequal across groups ($\delta^*_A \neq \delta^*_B$). The constraint arises because FPNED requires equal error rates (FPR/FNR) across groups, while AP demands equal accuracy, conditions that cannot hold simultaneously under unequal base rates without trivializing the classifier (\eg random predictions).

To understand why EO is not a desirable metric for fairness in target-group detection, consider a scenario with two marginalized demographic groups, \textit{Group-A} and \textit{Group-B}, that differ in how frequently they are targeted. When optimizing for EO, the model aligns true and false positive rates across both groups, ignoring the label imbalance. However, this alignment often requires redistributing errors, which disproportionately impacts the statistical minority group. This is not ethically desirable for several reasons. First, ``target group" refers to the group that a post is ``directed at or about'' \cite{sachdeva2022assessing}, which is independent of the downstream task (\eg whether the post is toxic). Being a frequent ``target" does not inherently mean a group warrants prioritization. Second, a group that constitutes a small minority in society may be statistical minority in the data (across both toxic and non-toxic posts) simply by virtue of its size. This does not mean it should be de-prioritized. 

\begin{table}[h]
    \centering
    \caption{Illustration of two marginalized groups with one (A) targeted more often than the other (B). Assume 200 total posts with {\em Group-A} having higher base rate. To optimize for EO (\textbf{Case I}), we set $TPR=0.80$ and $FPR=0.30$ across both groups. Alternatively, to optimize for AP (\textbf{Case II}), we set similar $Acc=0.77$ across both groups.}
    \begin{tabular}{l|cc|cccc|ccc} \toprule
        Group & Targeted & Not Targeted & TP & FN & FP & TN & Acc & TPR & FPR \\ \midrule
        \multicolumn{10}{c}{I - Optimizing for Equalized Odds} \\ \midrule 
        \textit{Group-A} & 100 & 100 & 80 & 20 & 30 & 70 & 0.75 & {\bf 0.80} & {\bf 0.30} \\
        \textit{Group-B} & 20 & 180 & 16 & 4 & 54 & 126 & 0.71 & {\bf 0.80} & {\bf 0.30} \\
        \midrule
        \multicolumn{10}{c}{II - Optimizing for Accuracy Parity} \\ \midrule
        \textit{Group-A} & 100 & 100 & 77 & 23 & 23 & 77 & {\bf 0.77} & 0.77 & 0.23 \\
        \textit{Group-B} & 20 & 180 & 15 & 5 & 41 & 139 & {\bf 0.77} & 0.75 & 0.23 \\ \bottomrule
    \end{tabular}
    \vspace{-0.75em}
    \label{tab:case-eo-acc}
\end{table}

To illustrate how EO leads to a statistical minority groups being disadvantaged, we present a synthetic example (\textbf{Table~\ref{tab:case-eo-acc}}), where both groups have $TPR=0.8$ and $FPR=0.3$. Despite identical error rates, \textit{group-B} exhibits lower accuracy $(0.71)$ due to its lower base rate and increased number of false positives. In contrast, optimizing for AP (Case II) equalizes overall accuracy across groups (0.77), reduces FPR for the \textit{group-B}, and maintains similar TPR/FNR values. Note that the increase in FPR that \textit{group-B} experiences under EO can lead to downstream harms. If the labels provided by such a target group detection module are used to train toxicity detection systems such as~\citet{gordon2022jury,fleisig2023majority,gupta2023same}, then these models would likely fail at learning group-specific toxicity patterns affecting \textit{group-B}, because the module often incorrectly flags this group as the target.

\section{Experimental Setup}

In this section, we describe the datasets used, our model architecture, the baseline loss, and two baseline fairness losses used for comparing performance of the $GAP_{multi}$ loss. Implementation details regarding setup, data pre-processing, hyperparameter tuning and related items can be found in \textbf{Appendix \ref{app:details}}.

\textit{Datasets}. We use two datasets for evaluation: the MHS corpus \cite{sachdeva2022assessing} and HateXplain \cite{mathew2021hatexplain}. 
MHS is available via the HuggingFace library \cite{huggingfaceDlab} and includes 135,556 posts sourced from YouTube, Twitter and Reddit, where each post is annotated for the demographic group(s) it targets. There are seven such groups: Asian, Black, Latinx, Middle-Eastern, Native-American, Pacific-Islander and White. We include posts targeting one or more groups, irrespective of toxicity labels (which we do not use). The Black community is the statistical majority ($\sim$22k posts), while Native American and Pacific Islander are statistical minorities ($\sim$2k posts). Details on the frequency with which different groups are targeted in the data, and the number of groups targeted by posts, are included in \textbf{Appendix Fig.}~\ref{fig:dems}. We do a random train-validation-test split of $70$-$10$-$20\%$ on the MHS corpus, since it does not come with pre-defined splits. 

HateXplain contains 57,687 posts sourced from Twitter and Gab. There are five racial groups---African, Arab, Asian, Caucasian, and Hispanic---and the African group is the majority group ($\sim$9k posts). HateXplain comes with its own $80$-$10$-$10\%$ predefined train-validation-test split. See Appendix \ref{app:details} for more details on the group splits across demographics and target cardinality, for both the MHS corpus and HateXplain datasets.

Incorporating data from multiple and diverse platforms is critical for validation of the different approaches compared in this work. The datasets provide multi-platform diversity, thereby allowing evaluation beyond platform-specific artifacts like slangs, structural format (\eg lengths allowed) or distributional shifts (\eg different communities and content having different prevalence across platforms). Both datasets contain targeted groups that are overly-represented (\eg African//Black) and under-represented (\eg Middle Eastern//Arab), allowing us to test the impact of sample size and class distribution on group fairness.
By ensuring our proposed approach performs consistently across these varied digital signatures and distributional shifts, we demonstrate its utility for adaptable solution across content analysis tasks. Moreover, by evaluating on two different datasets, we ensure that our results do not hinge on idiosyncrasies of how any one  particular dataset was sampled or annotated \cite{geva2019we}.


\textit{Model Architecture and Baseline Loss}. We ran experiments using DistilBERT \citep{sanh2019distilbert} and RoBERTa-base \cite{liu2019roberta}, using frozen weights to extract feature representations for posts. While RoBERTa has a richer representation than DistilBERT, in our case it did not yield any significant gain in performance, but it did add to runtime because of its larger size. Consequently, we discuss only DistilBERT results in the main body. See \textbf{Appendix \ref{app:perf}} for RoBERTa results. 

\begin{table*}[ht]
    \centering
    \caption{Balanced Accuracy (BA) 
    results on MHS for each loss function (OE and CLA baselines \vs $GAP_{multi}$) for the 7 demographic groups averaged over five runs. We also color which group exhibits the \textcolor{blue}{maximum} and \textcolor{red}{minimum} BA values for each loss over the 7 groups. The difference between this maximum \vs minimum is shown in the Max. Diff.\ column (for BA, we want to minimize this difference). $GAP_{multi}$ achieves a lower maximum difference (Max. Diff. = 5.54) than either baseline loss functions, evident from the visualization in \textit{Fig.~\ref{fig:ba7group}}. $GAP_{multi}$ also achieves the highest (macro) average BA across groups (Avg. BA = 81.97), and the best BA for most of the groups (\#Best BA column = 5/7). We also report the standard deviation of Max. Diff and Avg. BA over five runs, with CLA and $GAP_{multi}$ having similar deviation. $GAP_{multi}$ achieves the best fairness performance while also retaining overall utility.}
    \vspace{-0.5em}
    \resizebox{\textwidth}{!}{
    \begin{tabular}{l|ccccccc|rrc}
    \toprule
    & \multicolumn{7}{c|}{\textbf{MHS Corpus - Balanced Accuracy (BA) ($\uparrow$)}} & & & \\ \midrule
    Loss & Asian & Black & Latinx & Middle Eastern & Native American & Pacific Islander & White & \textbf{Max. Diff. ($\downarrow$)} & \textbf{Avg. BA ($\uparrow$)} & \textbf{\#Best BA ($\uparrow$)} \\ \midrule
    OE & 80.31 & \textbf{\textcolor{blue}{86.91}} & 81.07 & \textbf{84.87} & \textcolor{red}{64.99} & 67.91 & 75.01 & 21.9$\pm$1.3 & 77.29$\pm$0.29 & 2/7 \\
    CLA & 82.51 & \textcolor{blue}{85.34} & 81.67 & 84.62 & \textcolor{red}{73.91} & 74.92 & 80.44 & 11.4$\pm$0.8 & 80.49$\pm$0.14 & 0/7 \\ \midrule
    $GAP_{multi}$ & \textbf{83.18} & \textcolor{blue}{83.86} & \textbf{83.47} & 83.42 & \textbf{78.95} & \textbf{\textcolor{red}{78.32}} & \textbf{82.58} & \textbf{5.5}$\pm$0.5 & \textbf{81.97}$\pm$0.13 & \textbf{5/7} \\ \bottomrule
    \end{tabular}
     }
    
    \label{tab:ba7group}
\end{table*}

\begin{figure}[h]
    \centering
    \includegraphics[width=0.75\linewidth]{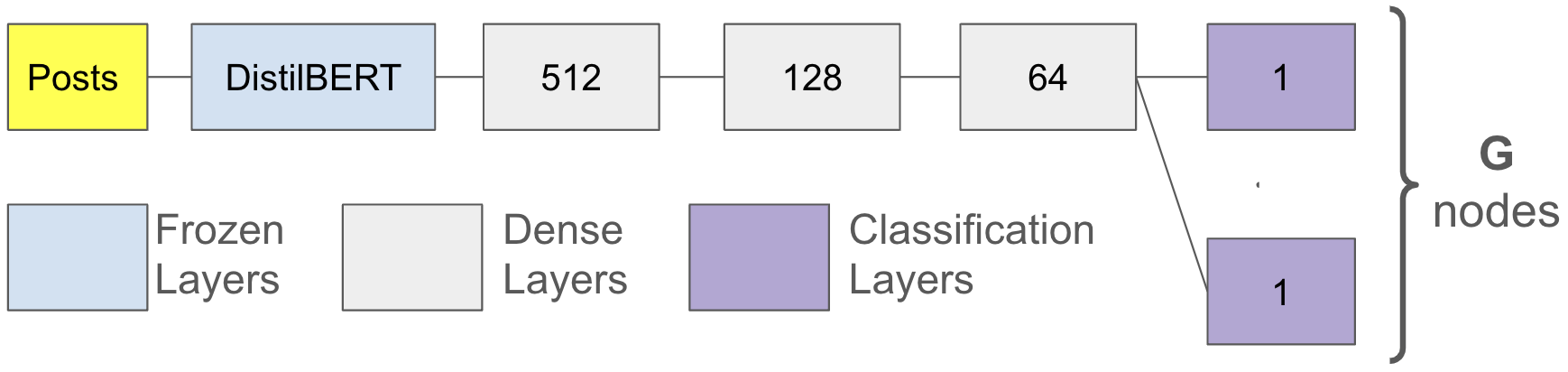}
    \vspace{-0.5em}
    \caption{Our multi-group target detection architecture. The model has shared parameters to learn both general and group-targeted language properties. The classification nodes each learn group-specific mappings.}
    \label{fig:arch}
    \vspace{-1.0em}
\end{figure}

This is followed by dense neuron layers with \textit{relu} activation, added biases and dropout rate of 0.1. 
The layers end in $|G|$ classification nodes, each with \textit{sigmoid} activation, no biases and 0.5 threshold (\textbf{Fig.~\ref{fig:arch}}). Given our multi-label binary classification setup, we use weighted Binary Cross Entropy (wBCE) as our baseline loss to optimize Overall Error (OE). 

\textit{Other Fairness Measures}.
We compare to CLAss-wise equal opportunity (CLA) \cite{shen2022optimising}, a differentiable fairness loss that
seeks to balance False Negative Rate (FNR) across groups \cite{chouldechova2017fair,hardt2016equality} by
minimizing error in absolute differences between error \wrt a label $(BCE(y))$ \vs error \wrt a label given the group 
$(BCE(y,g))$. We replace BCE with wBCE in Eq.~\ref{eq:cla} to compare fairly under label imbalance. 
\begin{align}
		CLA = BCE + \lambda \cdot \sum_{y \in N} \sum_{g \in G} |BCE(y,g) - BCE(y)| \label{eq:cla}
\end{align}



\noindent We also compare against another differentiable loss: ADV \cite{xia2020demoting}. Since ADV only supports a 2-group setting, we present its results in \textbf{Appendix \ref{app:2group}} on the MHS corpus.

\paragraph{Evaluation Measures} For model evaluation, we report Balanced Accuracy (BA) per group, Avg. BA over groups, Hamming Loss and macro-level Precision, Recall, and F1. These measures present a comprehensive view of model performance across groups  
that treats group equally regardless of their size. All the above mentioned measures are defined in \textbf{Appendix \ref{app:perf}}.

\section{Results} \label{sec:results}

Results show that $GAP_{multi}$ not only achieves better group-fairness,
but also maintains competitive overall accuracy. This highlights the effectiveness of $GAP_{multi}$ in addressing the unique challenges of multi-group target detection. 
All reported results in the main text are for the 7-group setting in MHS Corpus and 5-group setting in HateXplain. Additional results are shown in \textbf{Appendix~\ref{app:perf}}. Performance on selected $\Mycomb[|G|]{2}$ pairs are in \textbf{Appendix~\ref{app:2group}}. 

\begin{figure}[ht]
    \centering
    \includegraphics[width=\linewidth]{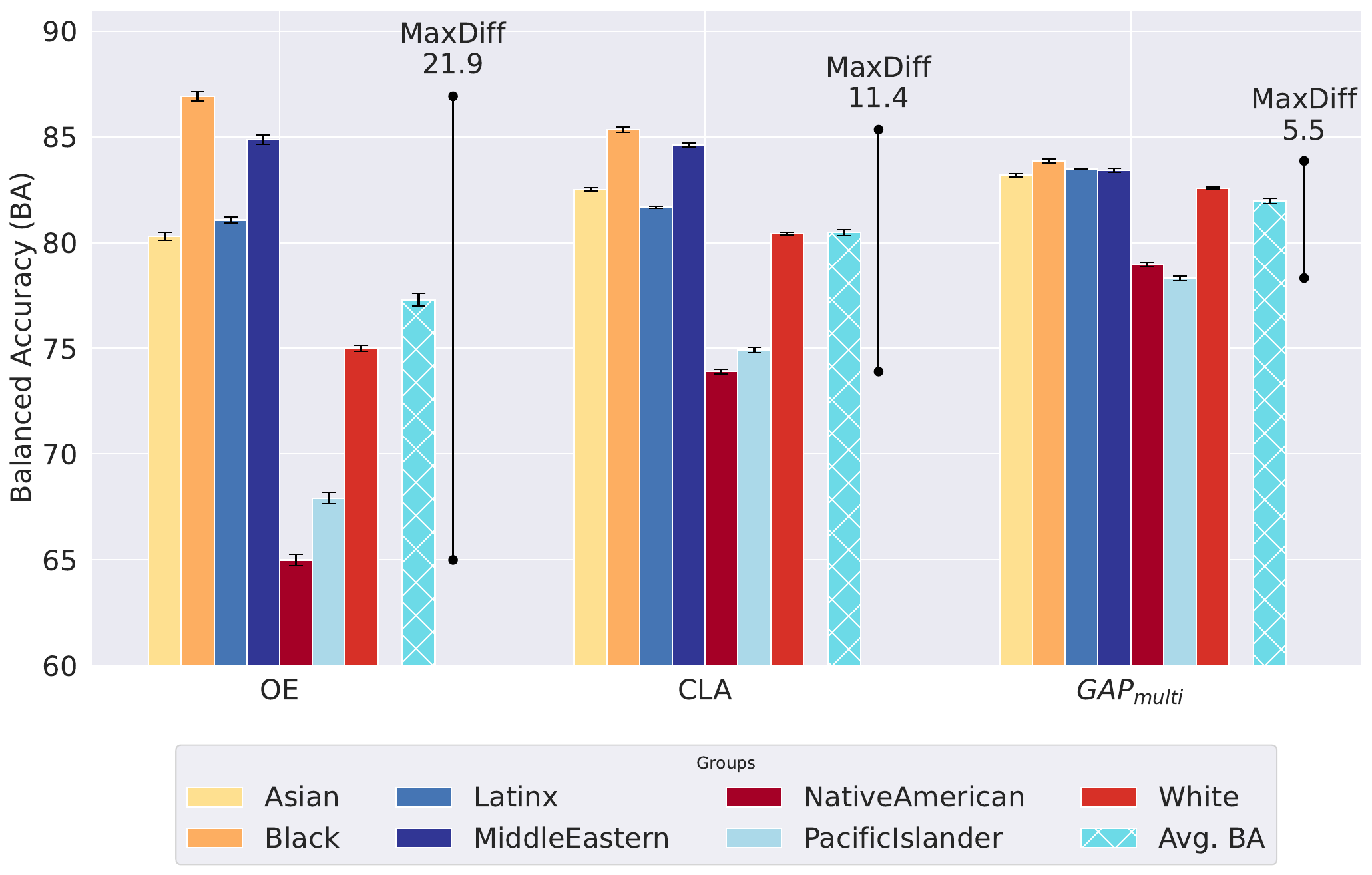}
    \vspace{-1.5em}
    \caption{Visualization of the BA values achieved by each loss over the 7 demographic groups in MHS corpus. The maximum difference (Max. Diff.) between the \textcolor{blue}{maximum} and \textcolor{red}{minimum} BA achieved for each loss across groups is also shown. 
    $GAP_{multi}$ performs best with lowest Max Diff. of 5.5, showcasing that it greatly reduces the disparities in performance across groups \vs other losses.}
    \label{fig:ba7group}
    \vspace{-0.75em}
\end{figure}

\textbf{Table~\ref{tab:ba7group}} shows results for the MHS Corpus dataset, with an equivalent visualization is shown in \textbf{Fig. \ref{fig:ba7group}}. We show BA values across groups averaged over five test set runs 
and compare the two baseline losses (OE and CLA) \vs $GAP_{multi}$. Recall that overall error (OE) minimizes $\overline{err}_{overall}$ without any fairness constraints. For the Black group, which is the statistical majority group in the dataset (see App. \textbf{Fig.~\ref{fig:dems}}), optimizing OE loss yields the highest BA (BA is generally higher for this group than others). Performance for the other groups---and in particular for the Native American group, which is a statistical minority---varies starkly across the different optimization strategies. The worst performing group for each loss (\ie having \textcolor{red}{minimum} BA values) is also seen to exhibit the best performance under $GAP_{multi}$. This means that $GAP_{multi}$ yields the best performance according to a min-max criteria \cite{diana2021minimax}. 

\begin{table*}[ht]
    \centering
    \caption{Balanced Accuracy (BA) results for each loss function (OE and CLA baselines \vs $GAP_{multi}$) for the 5 demographic groups on test data for HateXplain. For each loss, we color which group exhibits the \textcolor{blue}{maximum} and \textcolor{red}{minimum} BA values achieved for that loss over the 5 groups, with the difference between this maximum \vs minimum shown in the Max. Diff.\ column. $GAP_{multi}$ achieves a lower maximum difference (Max. Diff. = 5.19) than either baseline loss functions. We also report the standard deviation of Max. Diff and Avg. BA over five runs. $GAP_{multi}$ achieves the best fairness performance while also retaining overall utility.}
    \begin{tabular}{l|ccccc|rrc}
    \toprule
    & \multicolumn{5}{c|}{\textbf{HateXplain - Balanced Accuracy (BA) ($\uparrow$)}} & & & \\ \midrule
    Loss & African & Arab & Asian & Caucasian & Hispanic & \textbf{Max. Diff. ($\downarrow$)} & \textbf{Avg. BA ($\uparrow$)} & \textbf{\#Best BA ($\uparrow$)} \\ \midrule
    OE & \textbf{76.83} & 73.39 & \textbf{\textcolor{blue}{81.75}} & 70.91 & \textcolor{red}{70.79} & 10.96$\pm$0.8 & 74.73$\pm$0.15 & 2/7 \\
    CLA & 75.92 & 74.80 & \textcolor{blue}{79.15} & 72.50 & \textcolor{red}{71.85} & 7.30$\pm$0.6 & \textbf{74.84}$\pm$0.18 & 0/7 \\ \midrule
    $GAP_{multi}$ & 74.29 & \textbf{75.61} & \textcolor{blue}{77.46} & \textbf{73.65} & \textbf{\textcolor{red}{72.27}} & \textbf{5.19}$\pm$0.6 & 74.66$\pm$0.12 & \textbf{3/7} \\ \bottomrule
    \end{tabular}
    \vspace{-0.75em}
    
    \label{tab:ba7group-hatexplain}
\end{table*}

\begin{figure*}[ht]
    \centering
    \includegraphics[width=\linewidth]{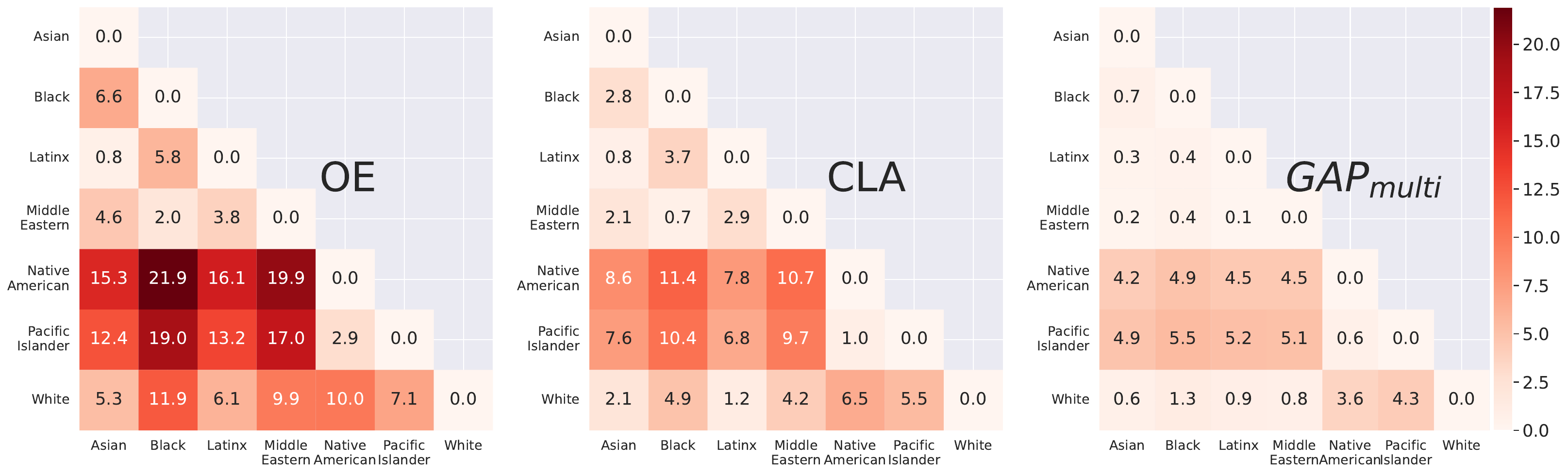}
    \vspace{-1.75em}
    \caption{Heatmap of pairwise absolute difference of BA across groups in test set (MHS corpus) as an indicator for bias and disparate impact. OE has the highest performance gap (Max Diff = 21.9) across groups as indicated by the extremes of color, not only across one group- pair but consistently across multiple group pairs. $GAP_{multi}$ has the least spread in pairwise error values (\textbf{Max. Diff. = 5.5}), evident from the flatness of color, indicating least disparate impact across groups.}
    \label{fig:pairwise-diff7group}
    \vspace{-0.75em}
\end{figure*}

We show the maximum difference (Max. Diff.) between the groups in \textbf{Table~\ref{tab:ba7group}}, which captures the gap in performance between worst \vs best performing groups. We see that solely optimizing for OE fails to account for variations in group performance, resulting in the highest difference values (\textit{Max. Diff. of 21.9}). In contrast, both $GAP_{multi}$ and CLA consider group performance alongside OE, leading to substantially lower Max. Diff. in comparison. Notably, $GAP_{multi}$ optimizes for balanced error rates across groups, exhibiting the smallest difference (\textit{Max. Diff. of 5.5}). \textbf{Table~\ref{tab:ba7group}} reports average BA (macro) obtained by each loss, with $GAP_{multi}$ performing best (\textbf{Avg. BA: 81.97}). We hypothesize that by incorporating group information in the loss during model training, both group-informed losses (CLA and $GAP_{multi}$) find better local optima \vs OE. The ``\#Best BA'' column reports for each loss function how many groups achieved the best BA for that loss. We see $GAP_{multi}$ performs best for five groups, while OE is best for the other 2 groups (\textit{Black, Middle Eastern}). This further shows that $GAP_{multi}$ does not prioritize optimizing the performance of one group over others, thereby being the best performing loss for most of the groups.

For the HateXplain dataset, the results presented in \textbf{Table~\ref{tab:ba7group-hatexplain}} show a consistent trend to our findings. The same experimental setup reveals that $GAP_{multi}$ again achieves the most balanced group performance, with the smallest Max. Diff. (5.19) between groups and similar overall utility (74.66). For this dataset, CLA achieves the highest Avg. BA (74.84), but both OE and $GAP_{multi}$ achieve statistically comparable performance. Meanwhile, $GAP_{multi}$ achieves a  statistically significant reduction in Max. Diff..  The consistent strong performance of the proposed approach across two distinct datasets with varying platforms, discourse style, and demographic composition shows that $GAP_{multi}$'s optimization for balanced group error is generalizable and not an artifact of a single data source. The comparative reduction in disparity across groups, alongside similar Avg. BA, shows the method's effectiveness as a fair and utility-preserving loss function for multi target-group identification task. The equivalent visualization is shown in \textbf{App Fig. \ref{fig:pairwise-diff7group-hatexplain}}.

To further illustrate performance disparities between groups, \textbf{Fig.~\ref{fig:pairwise-diff7group}} displays pairwise absolute differences of Balanced Accuracy (BA) across various groups ($|\textit{left - bottom}|$) in the MHS corpus. Higher heatmap values indicate classifier bias, revealing consistent inequities, particularly for overall error (OE) rates, where the (\textit{Black, Native American}) pair has the largest Max. Diff. (\textbf{21.9}). 
Two key observations can be drawn. 
First, the statistical majority \textit{Black} group dominates performance over the statistical minorities \textit{Native American} group. Second, disparities persist even for group pairs with similar prevalence (\eg \textit{Middle Eastern} \vs \textit{White}, BA gap: 9.9) because one group may simply be easier to identify than the other. While CLA (optimized for FNR) improves over OE, its misaligned objective (minimizing false negatives, rather than equitable performance), results in a 
heatmap in-between that of OE and $GAP_{multi}$. In contrast, the $GAP_{multi}$ loss, explicitly designed for balanced group performance, 
shows substantially fewer extremes (\textbf{Max. Diff.: 5.5}), with smoother heatmap transitions indicating equitable classifier outcomes.
Note that the extreme Max. Diff. values presented in Table~\ref{tab:ba7group} are reflected as extreme values in these heatmaps for specific group pairs. 
Similar trends can be observed for the pairwise absolute differences of BA heatmaps on HateXplain (\textbf{App. Fig. \ref{fig:pairwise-diff7group-hatexplain}}), where $GAP_{multi}$ has smoother heatmap transitions compared to the other two losses.

\begin{table}[h]
    \centering
    \caption{Summary statistics of other evaluation measures for the MHS corpus and HateXplain. Since CLA strictly optimizes for minimizing FNR, it indeed achieves the highest Recall (1 - FNR). However, this comes at the cost of losing out on Precision. Since $GAP_{multi}$ maximizes the average of TPR and TNR, it performs best both in terms of Precision and F1 scores across three losses.}
    \begin{tabular}{l|ccc} \toprule
    \multicolumn{1}{c}{} & \multicolumn{3}{c}{\textbf{MHS Corpus}} \\ \midrule
    Loss & $Prc_{macro}$ ($\uparrow$) & $Rec_{macro}$ ($\uparrow$) & $F1_{macro}$ ($\uparrow$) \\ \midrule
    OE & 0.7083 & 0.5808 & 0.6383 \\
    CLA & 0.5418 &\textbf{ 0.7143} & 0.6162 \\ \midrule
    $GAP_{multi}$ & \textbf{0.7854} & 0.6837 & \textbf{0.7310} \\ \midrule
    \multicolumn{1}{c}{} & \multicolumn{3}{c}{\textbf{HateXplain}} \\ \midrule
    Loss & $Prc_{macro}$ ($\uparrow$) & $Rec_{macro}$ ($\uparrow$) & $F1_{macro}$ ($\uparrow$) \\ \midrule
    OE & 0.7379 & 0.7148 & 0.7261\\
    CLA & 0.5789 & \textbf{0.7492} & 0.6531 \\ \midrule
    $GAP_{multi}$ & \textbf{0.7519} & 0.7365 & \textbf{0.7441} \\ \bottomrule
    \end{tabular}
    
    \label{tab:other_measures}
   \vspace{-0.75em}
\end{table}


\textbf{Table~\ref{tab:other_measures}} shows summary statistics for Precision, Recall and F1 for multi-label classification setup at the macro level on the 7-group MHS corpus and 5-group HateXplain dataset. The CLA approach, characterized by its emphasis on minimizing the False Negative Rate (FNR), yields the highest Recall. However, this optimization strategy comes at the expense of Precision and F1, both being lower than OE across the two datasets. Our proposed $GAP_{multi}$ loss, designed to jointly maximize the average of True Positive Rate (TPR) and True Negative Rate (TNR), is seen to be the best performing loss function in terms of both Precision and F1 score across the evaluated losses.


\textbf{Table~\ref{tab:hl_errors}} reports Hamming Loss (HL), 
the average fraction of misclassified labels for the MHS corpus and HateXplain. While CLA (HL: 12.10, 7.22) prioritizes minimizing FNR (which inherently involves asymmetry), at times it disproportionately optimizes for this aspect at the expense of other performance metrics. Consequently, CLA may exhibit poorer HL compared to OE (HL: 7.65, 6.47). In contrast, $GAP_{multi}$ maintains symmetry, achieving a balanced trade-off while maximizing the average of TPR and TNR. We see $GAP_{multi}$ consistently achieves the lowest HL: \textbf{6.85, 5.89}. 

\begin{table}[h]
    \centering
    \caption{HL values across losses for test data in the MHS corpus and HateXplain. \textbf{Lower} values are \textbf{better}. $GAP_{multi}$ optimizes for the average of TPR and TNR, thereby achieving lowest values, indicating better classifier performance.}
    \begin{tabular}{l|ccc|ccc} \toprule
    \multicolumn{1}{c}{} & \multicolumn{3}{c|}{\textbf{MHS Corpus}} & \multicolumn{3}{c}{\textbf{HateXplain}} \\ \midrule
         & OE & CLA & $GAP_{multi}$ & OE & CLA & $GAP_{multi}$ \\ \midrule
         \textbf{Hamming Loss (HL) \% ($\downarrow$)} & 7.65 & 12.10 & \textbf{6.85} & 6.47 & 7.22  & \textbf{5.89} \\ \bottomrule
    \end{tabular}
    \vspace{-0.75em}
    
    \label{tab:hl_errors}
\end{table}

\textbf{Runtime Performance} in \textbf{Table}~\ref{tab:runtime} shows results over the training dataset of 36k posts in MHS corpus, where we report the average time per epoch, number of epochs till convergence, total runtime, and the extra time ($\Delta$) for losses compared to OE. Since OE optimizes for weighted Binary Cross Entropy (wBCE), it takes the least time per epoch and the least number of epochs to converge. $GAP_{multi}$ takes all the steps for computing the overall loss in addition to calculating $\Mycomb[|G|]2$ pairwise losses and the balanced loss. Hence, $GAP_{multi}$ takes additional compute time for solving its intended optimization. The same argument holds true for CLA, since all of them are regularized single objective variants. Although $GAP_{multi}$ does the $\Mycomb[|G|]2$ extra computation, hence the extra runtime ($\Delta$: 1167s), it is not that significant (extra 9s per epoch) \wrt to OE, while gaining much more in terms of optimization improvement, because the pairwise error computations are all done in parallel. While OE and $GAP_{multi}$ operate on smooth losses their convergence epoch is relatively fast ($\sim$21, $\sim$27). CLA uses a 1-norm loss (Eq.~\ref{eq:cla}), hence the empirical loss surface is not as smooth as the previous two. As such, it is observable that CLA on average takes more epochs ($\sim$41) to converge with a higher $\Delta$.

\begin{table}[h]
    \centering
    \caption{Loss Runtime Analysis. While OE takes the least time, $GAP_{multi}$ gains more in optimizing performance across groups for an extra of $\sim 9s$ per epoch. The smoothness of $GAP_{multi}$ loss (27 epoch) also allows faster convergence compared to CLA (41 epoch).}
    \begin{tabular}{l|cccc}
        \toprule
         & \begin{tabular}[c]{@{}c@{}}\textbf{Avg. Time}\\ \textbf{Per Epoch (s)}\end{tabular} & \begin{tabular}[c]{@{}c@{}}\textbf{Epochs till}\\ \textbf{Convergence}\end{tabular} & \begin{tabular}[c]{@{}c@{}}\textbf{Runtime}\\ \textbf{Total (s)}\end{tabular} & \begin{tabular}[c]{@{}c@{}}$\Delta$ (s)\\ \wrt \textbf{OE}\end{tabular} \\
        \midrule
        OE & 154 & 21 & 3234 & 0 \\
        CLA & 158 & 41 & 6478 & 3244 \\ \midrule
        $GAP_{multi}$ & 163 & 27 & 4401 & 1167 \\
        \bottomrule
    \end{tabular}
    \vspace{-0.75em}
    
    \label{tab:runtime}
\end{table}

\begin{figure}
    \centering
    \includegraphics[width=0.8\linewidth]{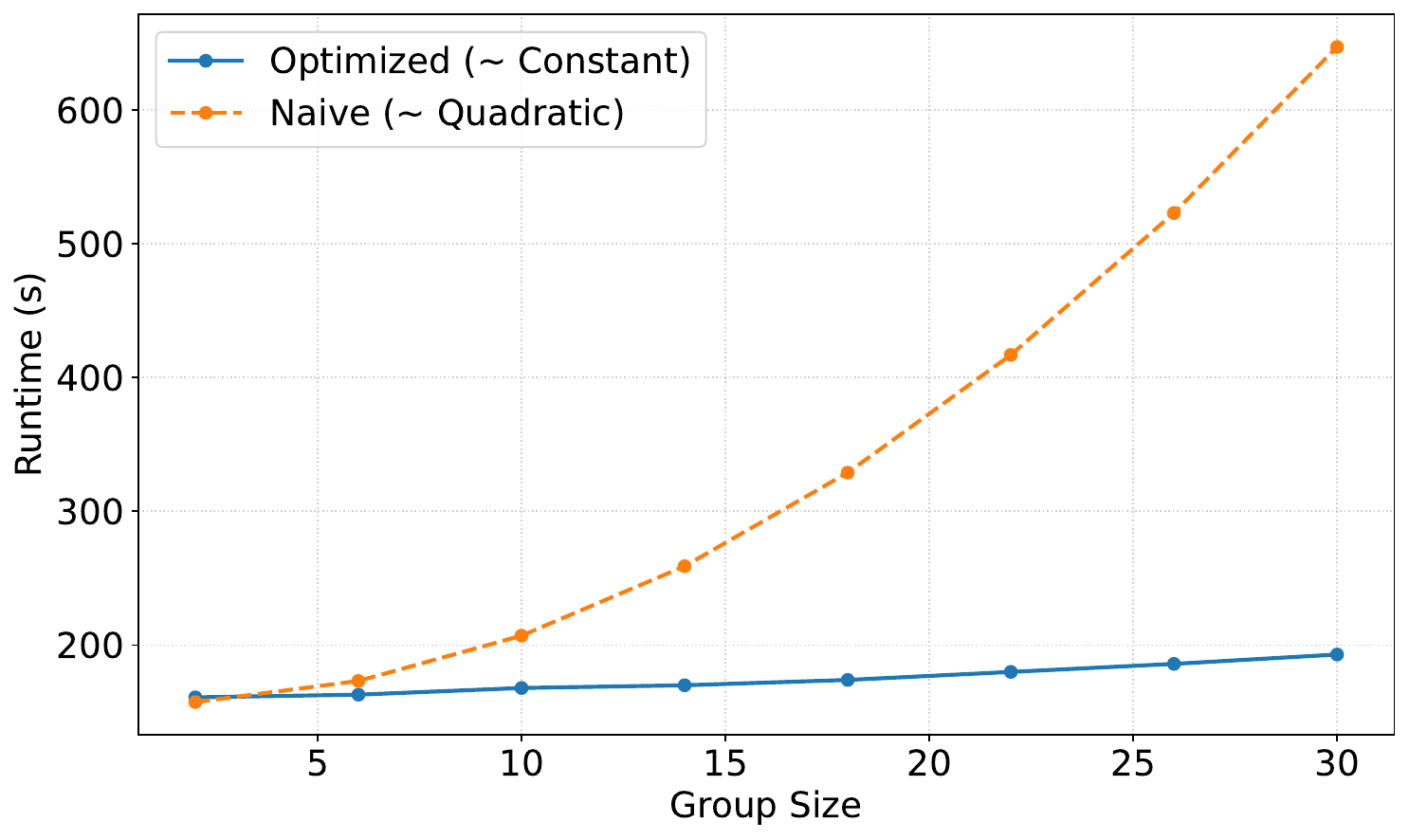}
    \vspace{-1.0em}
    \caption{Our $GAP_{multi}$ formulation enables parallel GPU computation of pairwise-group errors, achieving constant $\mathcal{O}(1)$ scaling complexity. Both the Optimized and Naive curves are for $GAP_{multi}$ loss with/-out the parallelism involved. The serial bottleneck in the Naive version leads to a quadratic scaling in runtime \wrt group size. For our implementation, the Optimized version has runtime per epoch with only  constant increase, even when scaling from 2 to 30 demographic groups.}
    \label{fig:time-scaling}
    \vspace{-1.5em}
\end{figure}

A unique advantage of our $GAP_{multi}$ formulation is that pairwise-group errors can be computed in parallel in GPU, thus having $\mathcal{O}(1)$ scaling \wrt group cardinality. For illustration, we report runtime per epoch with an increasing number of groups from 2-30 on a synthetic dataset, with each group having 22k samples. \textbf{Fig. \ref{fig:time-scaling}} shows that even at \#group:30 there is only constant increase in runtime per epoch compared to \#group:2 or 6. This is because the calculation of overall error $\overline{err}_{overall}$ is carried out independently and in parallel to the inter-group disparity $dev_{overall}$. The naive version has quadratic scaling resulting in runtime blowup \wrt group size, due to the serial computation bottleneck.

\section{Discussion}

Fairness measures must be carefully selected by practitioners to align with the requirements of the task, context, and stakeholders involved. In multi-group target detection, where posts may simultaneously target multiple groups, the symmetric error costs---symmetric meaning that all groups deserve equal protection---motivate a fairness measure that enforces equitable treatment across all groups. We thus adopted Accuracy Parity (AP) as the fairness criterion, since it directly addresses disparities in group-level predictive accuracy. AP measures the extent to which no group, particularly those with smaller representation or lower base rates, experiences disproportionate harm. This is accomplished using the $GAP_{multi}$ loss function, which equivalently minimizes for AP. The formulation enforces that fairness is incorporated directly into the model training process, enabling balanced performance without requiring additional pre- or post-processing steps. 

Ensuring fairness across a growing number of demographic groups $G$ poses computational challenges, particularly for methods relying on sequential global metrics like deviations from mean. In contrast, our $GAP_{multi}$ formulation leverages a pairwise structure, computing error deviations between all group pairs 
$\Mycomb[|G|]{2}$ independently. This design enables parallelization, integrating seamlessly with modern deep learning pipelines and exploiting GPU acceleration for efficiency. Despite the quadratic growth in terms of $|G|$, the $GAP_{multi}$ loss retains its convergence properties, ensuring stable optimization even for high group cardinalities, making it robust for real-world multi-group scenarios.


For the 7-group setting in Table~\ref{tab:ba7group}, we see that by optimizing for group-related errors alongside overall error, both $GAP_{multi}$ and CLA driven classifiers achieve better overall performance (in terms of Avg. BA) in addition to their intended group-specific objective.
Given our multi-label target-group classification where the group identifiers serve as the labels, we hypothesize that the group indicator gives an extra feature dimension. This likely enables the model to learn something more about the data than it could without the group-specific information. By incorporating group-associated indicators, both $GAP_{multi}$ and CLA have a modified surface compared to OE, allowing convergence to a better optima. We see this pattern emerging in some of the 2-group setting: \eg the \textit{Latinx} \vs \textit{Middle Eastern} group-pair in Appendix \ref{app:2group}. 

Crucially, $GAP_{multi}$'s symmetric loss design yields two key advantages over CLA: 1. It avoids imbalance-prone 1-norm penalization, leading to smoother convergence (see Table \ref{tab:runtime} in Appendix). 2. Its pairwise group-error minimization promotes equitable performance across all labels, which doesn't risk over-prioritizing certain groups. In the experiments, and aligned with our theoretical proof characterizing the risks of optimizing for EO instead of AP, we find that CLA yields worse performance for statistical minorities, such as Native Americans and Pacific Islanders. Meanwhile, $GAP_{multi}$ minimizes gaps across all groups.

{\em Model Multiplicity} \cite{black2022model} refers to the phenomenon wherein several distinct machine learning models achieve nearly identical overall utility, such as accuracy, yet produce divergent predictions for individual data points. This is crucial with target-group detection, where classifiers trained with CLA and $GAP_{multi}$ can exhibit similar overall accuracy but greatly diverge in their group-level performances. This divergence reveals that model selection based solely on overall utility can mask severe disparities in how models treat different demographic groups. In this setting, AP serves as a fairness diagnostic providing useful insights into the degree of bias across groups. $GAP_{multi}$ achieves the smallest pairwise performance gap across groups, thereby promoting a more equitable distribution of utility.
 
\section{Conclusion}

We have motivated and proposed a framework for \textit{fairness-aware multi-group target detection}, focusing on realistic scenarios where social media posts may target multiple demographic groups simultaneously. Given the symmetric harm of detection errors in this setting, we adopted Accuracy Parity (AP) as a fairness criteria and proposed the $GAP_{multi}$ loss function to minimize bias in a multi-group setting. We jointly optimize for both group-fairness and utility, aligning training dynamics with evaluation measures in a robust and scalable manner. We also present theoretical analysis showing the incompatibility between AP and Equalized Odds (EO). Our empirical results show consistently reduced bias across groups while retaining overall utility on {\em fairness-aware multi-group target detection} and thereby the proposed loss function contributes towards developing safer and more inclusive online spaces.

\section{Ethical Considerations Statement}



In safety-critical domains, such as toxicity detection and content moderation, the ability to identify specific groups serve a vital protective function by shielding vulnerable populations from harm and delivering a clear societal value that justifies its use. Conversely, in contexts like targeted advertising, the same technical capability could enable profiling or manipulation without stringent oversight, posing significant ethical risks. Researchers and practitioners should acknowledge potential risks alongside potential benefits of work in this area, and to make every effort to ensure that potential benefits outweigh potential risks. To this end, we explicitly recommend that practical deployments prioritize protective applications, incorporate transparent ethical guidelines, and implement rigorous impact assessments. Moreover, users should always be informed of when and why such target-group detection systems are deployed. This approach strives to ensure and support our goal that the proposed tool aligns with societal values and mitigates the risk of exacerbating existing inequalities.
Our experiments used two existing, public datasets, the MHS corpus \cite{sachdeva2022assessing} and HateXplain \cite{mathew2021hatexplain}. For MHS, its authors note that their data annotation process was approved by UC Berkeley's IRB. Their annotators were allowed to omit any demographic information, and all data samples were also anonymized to protect annotator privacy. For HateXplain, its authors note that posts were anonymized by replacing the usernames with <user> token. Our own experiments use only the target's demographic-group flag and we do not infer any other user/annotator sensitive information.

\section*{Acknowledgments} 
We thank the data annotators who made this work possible, as well as our anonymous reviewers for their valuable feedback. This research was supported in part by Amazon, Wipro, the Knight Foundation, the Micron Foundation, and by Good Systems\footnote{\url{https://goodsystems.utexas.edu/}}, a UT Austin Grand Challenge to develop responsible AI technologies. The statements made herein are solely the opinions of the authors and do not reflect the views of the sponsoring agencies.

\bibliographystyle{ACM-Reference-Format}
\bibliography{References}

\appendix

\section{Impossibility Proofs} \label{app:proof}

Assume two groups $A$ and $B$. Let the number of positive and negative examples in group $A$ be $P_A$ and $N_A$ respectively. Similarly, $P_B$ and $N_B$ for group $B$. Proportion of positive to negative examples in a group, \ie $\delta_A=\frac{P_A}{N_A}$ and $\delta_B=\frac{P_B}{N_B}$.

Since we are satisfying Equalized Odds, we have:
\begin{align*}
    TPR_A = TPR_B = TPR \\
    FPR_B = FPR_B = FPR
\end{align*}

\noindent \textbf{Proof of Theorem 1.}

From basic definitions, we know the following:
\begin{align*}
    \textrm{True Positive Rate (TPR)} &= \frac{TP}{TP+FN} \\
    \textrm{False Positive Rate (FPR)} &= \frac{FP}{FP+TN} \\
    \textrm{Accuracy (Acc)} &= \frac{TP+TN}{TP+TN+FP+FN} \\
    &= \frac{TP+TN}{P+N}
\end{align*}

For Group A and Group B, we have the following:

\begin{minipage}{0.45 \linewidth}
\begin{align*}
    TP_A &= TPR * P_A \\
    FN_A &= P_A - TP_A = P_A*(1-TPR) \\
    FP_A &= FPR * N_A \\
    TN_A &= N_A - FP_A = N_A*(1-FPR) \\
    \therefore Acc_A &= \frac{TP_A+TN_A}{P_A+N_A} \\ &= \frac{TPR*P_A+N_A*(1-FPR)}{P_A+N_A} \numberthis \label{eq:accga}
\end{align*}
\end{minipage} \hfill
\begin{minipage}{0.45 \linewidth}

\begin{align*}
    TP_B &= TPR * P_B \\
    FN_B &= P_B - TP_B = P_B*(1-TPR) \\
    FP_B &= FPR * N_B \\
    TN_B &= N_B - FP_B = N_B*(1-FPR) \\
    \therefore Acc_B &= \frac{TP_B+TN_B}{P_B+N_B} \\ &= \frac{TPR*P_B+N_B*(1-FPR)}{P_B+N_B} \numberthis \label{eq:accgb}
\end{align*}
\end{minipage}

\begin{remark}
    A fairness problem aiming to simultaneously satisfy Equalized Odds and Accuracy is only feasible when the base rates are equal across groups.
\end{remark}

This remark extends \citet{kleinberg2016inherent}, showing that overall accuracy and EO (group-fairness) are at odds with each other for most practical cases when base rates differ across demographic groups ($\delta^*_A \neq \delta^*_B$), \sut one needs to be sacrificed for the gain of the other. Enforcing EO may thus come at the cost of significant accuracy degradation, particularly in imbalanced datasets. Hence, we can observe that group accuracies are directly dependent on the base rates of each group. Therefore, enforcing EO on that would cause one or both groups to deviate from  their optimal accuracy levels. For example if $\delta_A > \delta_B$, then to equalize TPR, group B must increase its TPR disproportionately, which decrease its accuracy levels.

Expanding on Eq.~\ref{eq:accga} and Eq.~\ref{eq:accgb}, we have the following two scenarios:

\textbf{Special Case 1:} Base rates across groups are equal, \ie $\frac{P_A}{N_A} = \frac{P_B}{N_B}$, $P_A=\alpha P_B, N_A=\alpha N_B$, $\forall \alpha \in (0, \infty)$.

Under this special setting of equalized base rates, Eq.~\ref{eq:accga} and Eq.~\ref{eq:accgb} becomes the same:
\begin{align*}
    Acc_A &= \frac{TPR*P_A+N_A*(1-FPR)}{P_A+N_A} \\
    &= \frac{TPR*\alpha P_B+\alpha N_B*(1-FPR)}{\alpha P_B+ \alpha N_B} \\
    &= \frac{\alpha (TPR*P_B+N_B*(1-FPR))}{\alpha (P_B+N_B)} \\
    &= Acc_B
\end{align*}

\textbf{Special Case 2:} Base rates are not equal, but $TPR$ and $FPR$ sum to one. $TPR+FPR=1$. Under this special setting of unequalized base rates, Eq.~\ref{eq:accga} and Eq.~\ref{eq:accgb} becomes the same. This is equivalent to random predictions by the classifier.

\begin{minipage}{0.45 \linewidth}
\begin{align*}
    Acc_A &= \frac{TPR*P_A+N_A*(1-FPR)}{P_A+N_A} \\
    &= \frac{TPR*P_A+N_A*TPR}{P_A+N_A} \\
    &= \frac{TPR (P_A+N_A)}{P_A+N_A} \\
    &= TPR
\end{align*}
\end{minipage} \hfill
\begin{minipage}{0.45 \linewidth}
\begin{align*}
    Acc_B &= \frac{TPR*P_B+N_B*(1-FPR)}{P_B+N_B} \\
    &= \frac{TPR*P_B+N_B*TPR}{P_B+N_B} \\
    &= \frac{TPR (P_B+N_B)}{P_B+N_B} \\
    &= TPR
\end{align*}
\end{minipage}

Case 1 demonstrates that AP and EO are mathematically compatible only when base rates are equal across groups, an assumption that rarely holds in practice, as most real-world datasets exhibit inherent demographic imbalances. Case 2 reveals a more troubling scenario: even with unequal base rates, simultaneous satisfaction of AP and EO forces the classifier into degenerate behavior, reducing it to random guessing (\ie predicting uniformly at chance level). This shows a fundamental tension, \ie strictly enforcing both criteria under realistic (imbalanced) conditions either violates feasibility (Case 1) or renders the model useless (Case 2).

\textbf{Proof of Theorem 2.} The False Positive and Negative Equality Difference (FPNED) \cite{chen2024hate} can be broken down into FPED and FNED. 
\begin{align*}
    FPED = \sum_G |FPR-FPR_g| \\ 
    FNED = \sum_G |FNR-FNR_g|
\end{align*}

Only when $FPR_A=FPR_B$ and $FNR_A=FNR_B$, does $FPED=0$ and $FNED=0$. Eq.~\ref{eq:accga} and Eq.~\ref{eq:accgb} can be rephrased as:

\begin{minipage}{0.45 \linewidth}
\begin{align*}
    Acc_A &= \frac{TP_A+TN_A}{P_A+N_A} \\ 
    &= \frac{(1-FNR)*P_A+(1-FPR)*N_A}{P_A+N_A}
\end{align*}
\end{minipage} \hfill 
\begin{minipage}{0.45 \linewidth}
\begin{align*}
    Acc_B &= \frac{TP_B+TN_B}{P_B+N_B} \\ 
    &= \frac{(1-FNR)*P_B+(1-FPR)*N_B}{P_B+N_B}
\end{align*}
\end{minipage}

For the accuracies to be same, we need $Acc_A = Acc_B$. Equating and rearranging the terms we have: 
\begin{align*}
   & P_A*(P_B+N_B) + N_A*(P_B+N_B) \\
   &= P_B*(P_A+N_A) + N_B*(P_A+N_A)
\end{align*}
which only holds when $\frac{P_A}{N_A} = \frac{P_B}{N_B}$, showing that AP and FPNED are only equivalent when base rates are equal.

\section{Experimental Details} \label{app:details}

\paragraph{Setup} Experiments use a Nvidia 2060 RTX Super 8GB GPU, Intel Core i7-9700F 3.0GHz 8-core CPU and 16GB DDR4 memory. We use Keras \citep{chollet2015} library on a Tensorflow 2.0 backend with Python 3.12 to train the networks in this paper. For optimization, we use AdaMax \citep{kingma2014adam} with parameters (\textit{lr}=0.001) and $1000$ steps per epoch. We also setup an EarlyStopping criteria in case the with a min\_delta of $1\textsc{e-}4$ and patience of 5. The random seed was set to 42 to avoid varaibility.

\paragraph{Optimizer} We choose AdaMax (infinity-norm variant of Adam) due to its faster convergence from fresh starts, \ie model training from random weights. It is also more stable than Adam in our case with sparse or unbalanced gradients. This makes tuning the learning rate less fragile, which is convenient when balancing $\lambda$ for fairness.

\paragraph{Regularization} We selected $\lambda$ using held-out validation. Early stopping triggered when either loss components reached plateaus (no significant improvement for 5 consecutive epochs) OR a minimum accuracy threshold (81\%$\pm$tolerance). Note that the 81\% threshold was selected based on post-priori analysis of different runs over various losses. Values of chosen $\lambda$ on both datasets are shown in \textbf{Fig. \ref{fig:tradeoff}} and \textbf{Fig. \ref{fig:tradeoff-hatexplain}}.

\begin{figure}[ht!]
    \centering
    \begin{subfigure}{0.49\linewidth}
        \centering
        \includegraphics[width=0.9\linewidth]{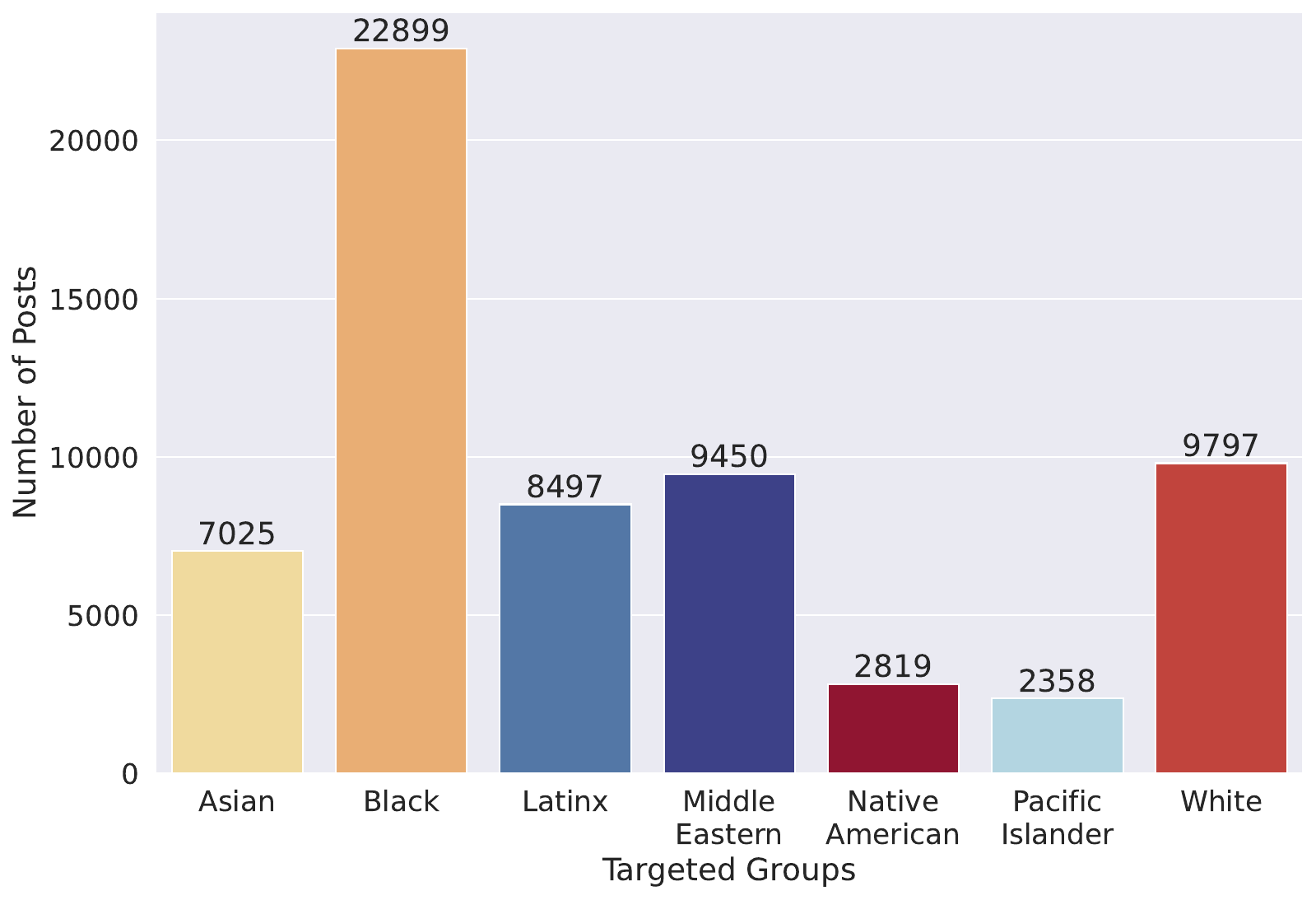}
        \vspace{-0.5em}
        \caption{Number of posts targeting each group.}
    \end{subfigure}
    \begin{subfigure}{0.49\linewidth}
        \centering
        \includegraphics[width=0.9\linewidth]{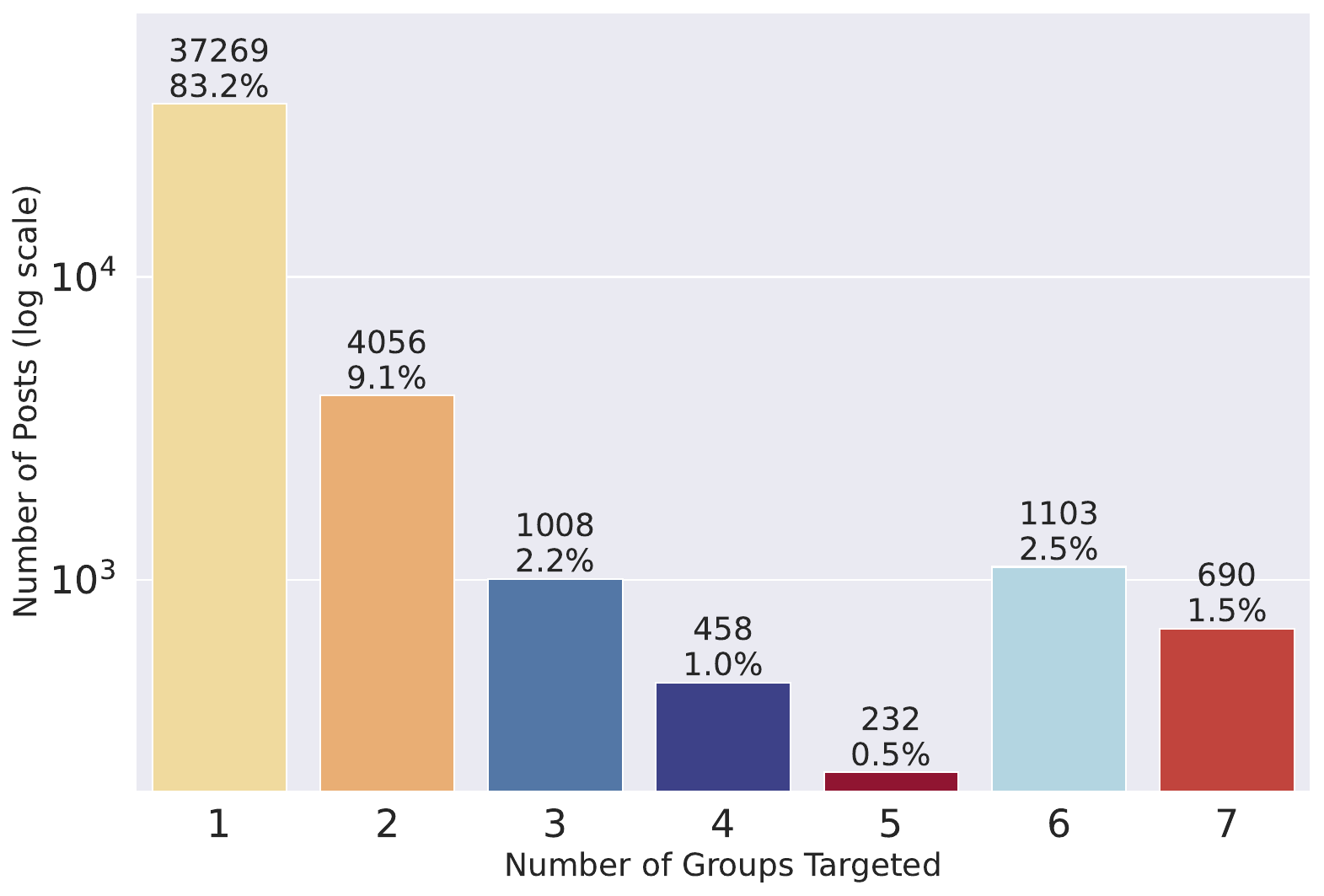}
        \vspace{-0.5em}
    \caption{Number of posts targeting multiple groups.}
    \end{subfigure}
    \vspace{-0.5em}
    \caption{Summary statistics of the MHS corpus \cite{sachdeva2022assessing} show the distribution of posts targeting demographic groups. The Black community is the statistical majority, while Native American and Pacific Islander are statistical minorities. Additionally, the dataset includes posts targeting multiple groups, reflecting its multi-group nature.}
    \label{fig:dems}
\end{figure}

\paragraph{Dataset} We use the MHS corpus \cite{sachdeva2022assessing} via the HuggingFace library \cite{huggingfaceDlab}. Its 135,556 posts and seven race groups: Asian, Black, Latinx, Middle-Eastern, Native-American, Pacific-Islander and White. \textbf{Fig.~\ref{fig:dems}(a)} shows how frequently each group is targeted. \textbf{Fig.~\ref{fig:dems}(b)} shows the number of groups targeted per post\footnote{For groups of size six and seven, posts target a broader range of demographic categories, often using inclusive terms like `non-white', as shown in Table~\ref{tab:task-setup}. This explains the higher frequency of such posts compared to those targeting fewer than five groups.}. 

\begin{figure}[ht!]
    \centering
    \vspace{-0.75em}
    \begin{subfigure}{0.49\linewidth}
        \centering
        \includegraphics[width=0.9\linewidth]{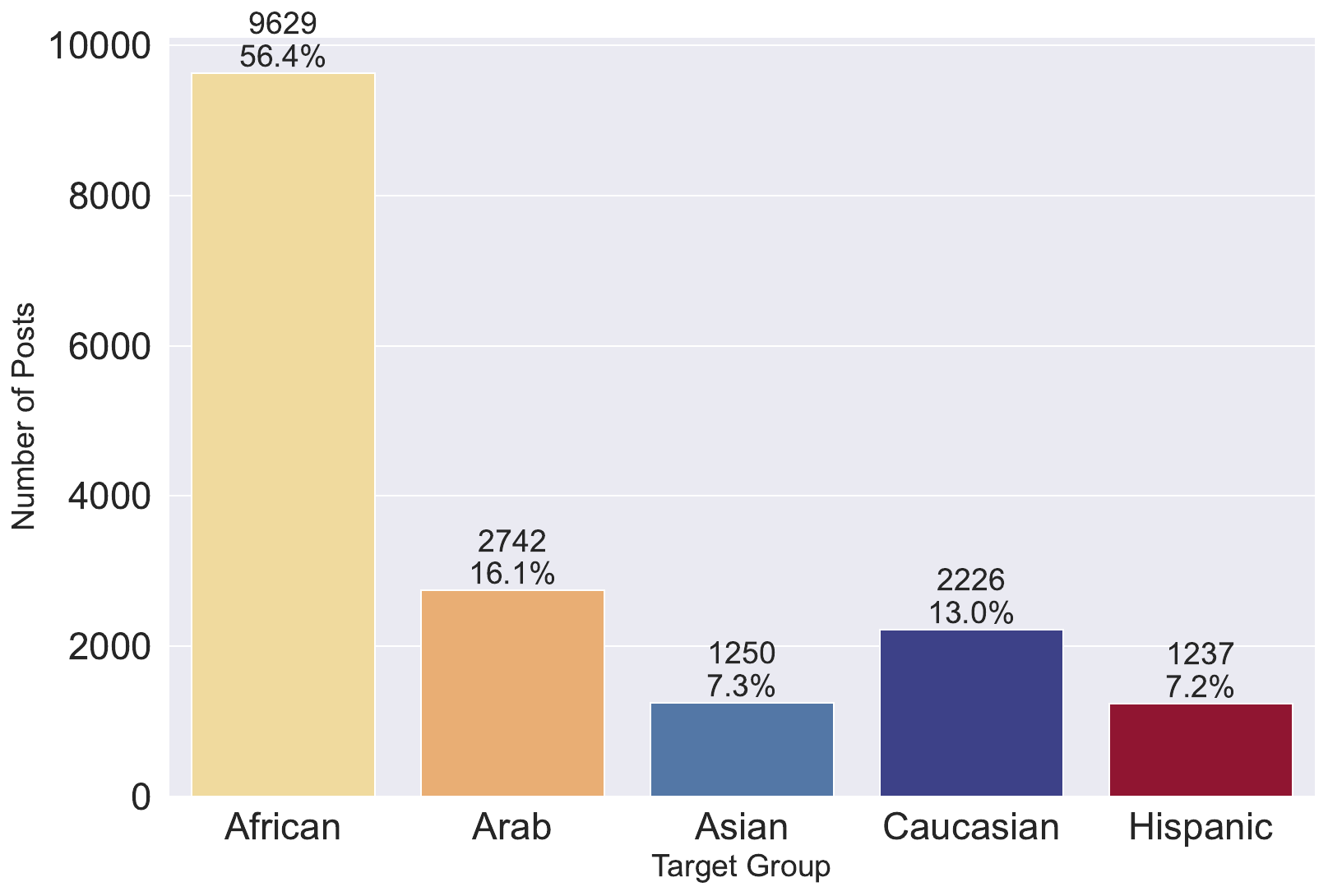}
        \vspace{-0.5em}
        \caption{Number of posts targeting each group.}
    \end{subfigure}
    \begin{subfigure}{0.49\linewidth}
        \centering
        \includegraphics[width=0.9\linewidth]{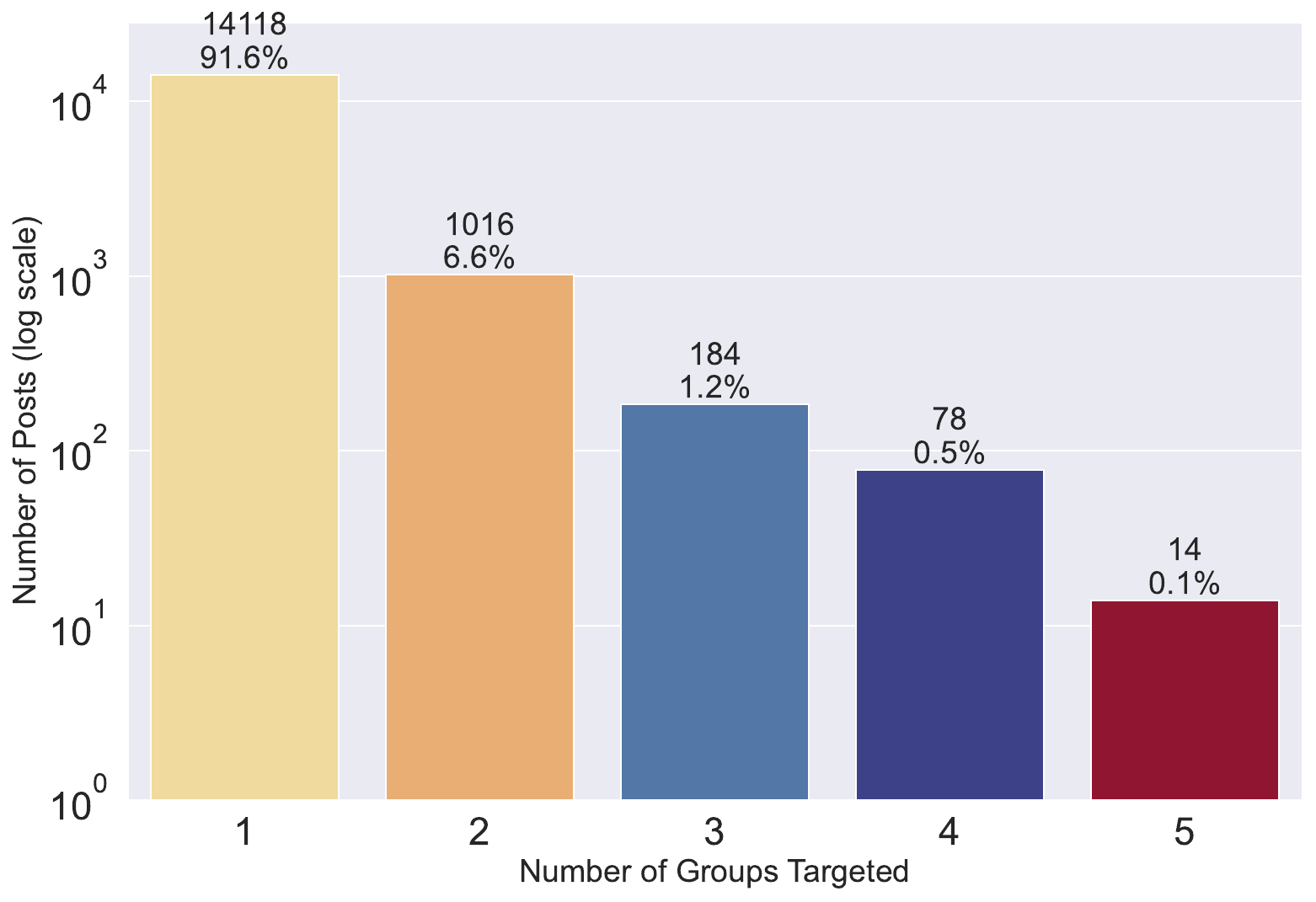}
        \vspace{-0.5em}
    \caption{Number of posts targeting multiple groups.}
    \end{subfigure}
    \vspace{-0.5em}
    \caption{Summary statistics of the HateXplain \cite{mathew2021hatexplain} show the distribution of posts targeting demographic groups. The African community is the statistical majority, while Native American and Pacific Islander are statistical minorities. Additionally, the dataset includes posts targeting multiple groups, reflecting its multi-group nature.}
    \vspace{-1.5em}
    \label{fig:dems-hatexplain}
\end{figure}

We also use the HateXplain \cite{mathew2021hatexplain} dataset containing 57,687 posts sourced from Twitter and Gab with the following demographic groups: African, Arab, Asian, Caucasian, and Hispanic, with African as the statistical majority group ($\sim$9k posts), while Asian and Hispanic are statistical minorities. 
\textbf{Fig.~\ref{fig:dems-hatexplain}(a)} shows how frequently each group is targeted. \textbf{Fig.~\ref{fig:dems-hatexplain}(b)} shows the number of groups targeted per post.

\paragraph{Dataset Pre-processing} From the set of all posts across both datasets, we filter ones that target at least one or more demographic group(s). In MHS Corpus this results in a dataset size of 44,816 posts. We perform a 70-10-20\% stratified split for training, validation and testing data. HateXplain's filtered posts has 15,410 samples, which comes with its own pre-defined 80-10-10\% training, validation and testing split. We further divide the data into clusters by number of targeted groups, for all cluster sizes that allowed for a SkLearn's \citep{scikit-learn} stratified sampling on the label split ratio.

\paragraph{Loss Function} Another fairness loss \cite{xia2020demoting} is an adversarial approach to demoting unfairness, which we denote as ADV. It seeks to provide false positive rate (FPR) balance \cite{chouldechova2017fair} across groups, otherwise known as \textit{predictive equality} (\textit{ibid.})
\begin{align}
		ADV = \beta \cdot BCE + (1-\beta) \cdot (adversary(y,g)-0.5) \label{eq:adv}
\end{align}

\section{Additional Experiments} \label{app:perf}

\subsection{Evaluation Measures}

\paragraph{Balanced Accuracy (BA)} 
Unlike standard accuracy, which can be misleading in the presence of label imbalance, 
Balanced Accuracy measures accuracy by averaging per-label performance, addressing dataset imbalance. By considering both the (\textit{TPR: true positive rate}) and (\textit{TNR: true negative rate}) of each label, BA effectively captures the model's ability to correctly classify instances across all labels, regardless of their prevalence.
\begin{align}
    BA = (TPR + TNR) / 2.0 \label{eq:ba}
\end{align}

\paragraph{Avg. BA} We report the macro averaged BAs as a summary statistic, to equally weigh performance across all groups.
The Avg. BA (macro) treats each group equally, ensuring that the classifier's performance is evaluated in a balanced manner across all demographic groups.
macro BA takes precedence over micro BA due to its emphasis on equal representation of groups and and fairness. Unlike micro BA, which heavily weights larger groups, macro BA 
\begin{align}
    Avg.\, BA = \frac{1}{G} \sum_{g =1}^G BA(g)
\end{align}

\paragraph{Hamming Loss (HL)} This is a widely used metric for assessing the performance of multi-label classifiers. For a dataset with $N$ instances and $G$ labels, the Hamming Loss measures the average fraction of incorrectly predicted labels across all instances, with $hamming(y_i(g),\hat{y}_i(g))$ as an indicator function of 1 if the $g$-th label for instance $i$ is incorrectly predicted and 0 otherwise. 
Specifically, it measures the average fraction of labels that are misclassified in comparison to the true label set. HL and Subset Accuracy Loss are comparable under small label cases \cite{wu2020multi}, hence we just report HL.
\begin{align}
    HL = \frac{1}{NG} \sum_{i=1}^{N} \sum_{g=1}^{G} hamming(y_i(g),\hat{y}_i(g)) \label{eq:hl}
\end{align}



\subsection{Results on MHS}

To assess loss function performance consistency across different text encoder representations, we also evaluate with RoBERTa-base \cite{liu2019roberta} instead of DistilBERT. RoBERTa-base did not yield any significant gain in utility, besides adding to runtime due to its larger size. Results shown in \textbf{Table~\ref{tab:ba7group-roberta}} validate that even with upgraded text encoder model, $GAP_{multi}$ achieves the minimum group disparity (Max. Diff.) performance among the compared losses.

\begin{table}[h]
    \centering
    \vspace{-1em}
    \caption{Evaluation of loss function on the MHS corpus using RoBERTa-base as the text encoder.}
    \resizebox{\textwidth}{!}{
    \begin{tabular}{l|ccccccc|rrc}
    \toprule
    & \multicolumn{7}{c|}{\textbf{MHS Corpus - Balanced Accuracy (BA) ($\uparrow$)}} & & & \\ \midrule
    Loss & Asian & Black & Latinx & Middle Eastern & Native American & Pacific Islander & White & \textbf{Max. Diff. ($\downarrow$)} & \textbf{Avg. BA ($\uparrow$)} & \textbf{\#Best BA ($\uparrow$)} \\ \midrule
    OE & 81.35 & \textbf{\textcolor{blue}{87.34}} & 81.65 & \textbf{85.12} & 67.85 & \textcolor{red}{67.53} & 79.76 & 19.81$\pm$0.9 & 78.66$\pm$0.26 & 2/7 \\
    CLA & 81.56 & \textcolor{blue}{85.36} & 81.76 & 83.82 & 75.33 & \textcolor{red}{70.43} & 80.12 & 14.93$\pm$0.6 & 79.77$\pm$0.18 & 0/7 \\ \midrule
    $GAP_{multi}$ & \textbf{81.82} & \textcolor{blue}{84.04} & \textbf{81.98} & 83.42 & \textbf{77.67} & \textbf{\textcolor{red}{77.58}} & \textbf{80.82} & \textbf{6.46}$\pm$0.6 & \textbf{81.04}$\pm$0.18 & \textbf{5/7} \\ \bottomrule
    \end{tabular}
     }
    \vspace{-0.75em}
    
    \label{tab:ba7group-roberta}
\end{table}

\begin{figure}[ht]
    \centering
    \includegraphics[width=0.6\linewidth]{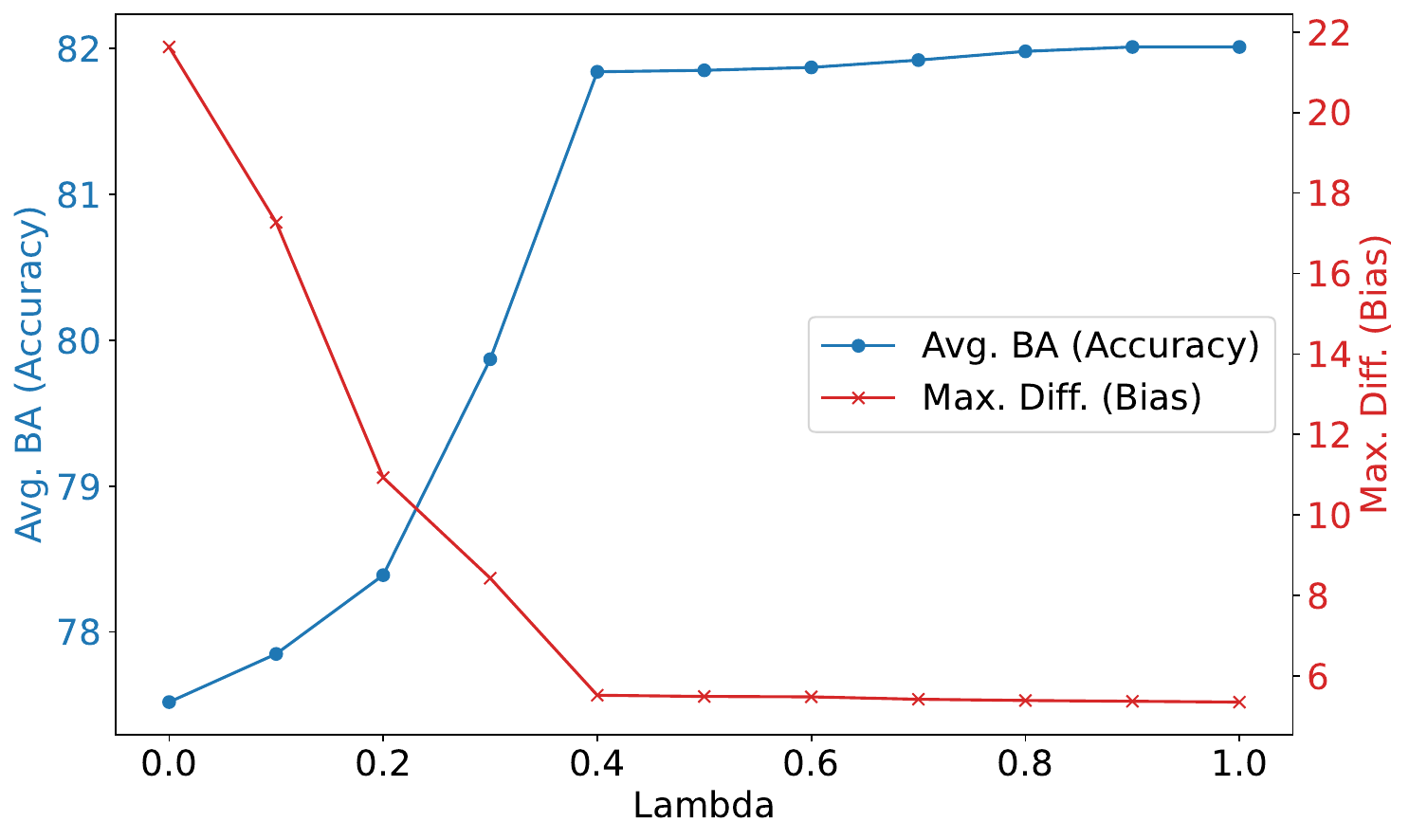}
    \vspace{-1.0em}
    \caption{Hyperparameter results showing variation of Avg. BA and Max. Diff. \vs $\lambda$ on the MHS corpus. The extra group dimension provides both CLA and $GAP_{multi}$ an extra dimension to optimize over, hence we empirically gain improvements in both vertical axes.}
    \label{fig:tradeoff}
    \vspace{-0.75em}
\end{figure}

For $\lambda$-tuning on MHS, see \textbf{Fig. \ref{fig:tradeoff}} for sensitivity analysis. The stopping criteria for this experiment was triggered at $\lambda$=0.4, validation accuracy stabilized at 81.84\%. Note that the trajectory of the curve goes beyond traditional optimization (Eq. \ref{eq:soo}) where with increasing fairness (lower bias), overall accuracy should be correspondingly decrease. However, as we discussed for MHS, both CLA and $GAP_{multi}$ have improvements in accuracy possibly due to the extra group dimension that is invoked during optimization. Hence for this specific data, we have empirically gained in both accuracy and group fairness with increasing $\lambda$. 

\subsection{Results on HateXplain}

\begin{figure}[ht]
    \centering
    \vspace{-0.5em}
    \includegraphics[width=0.9\linewidth]{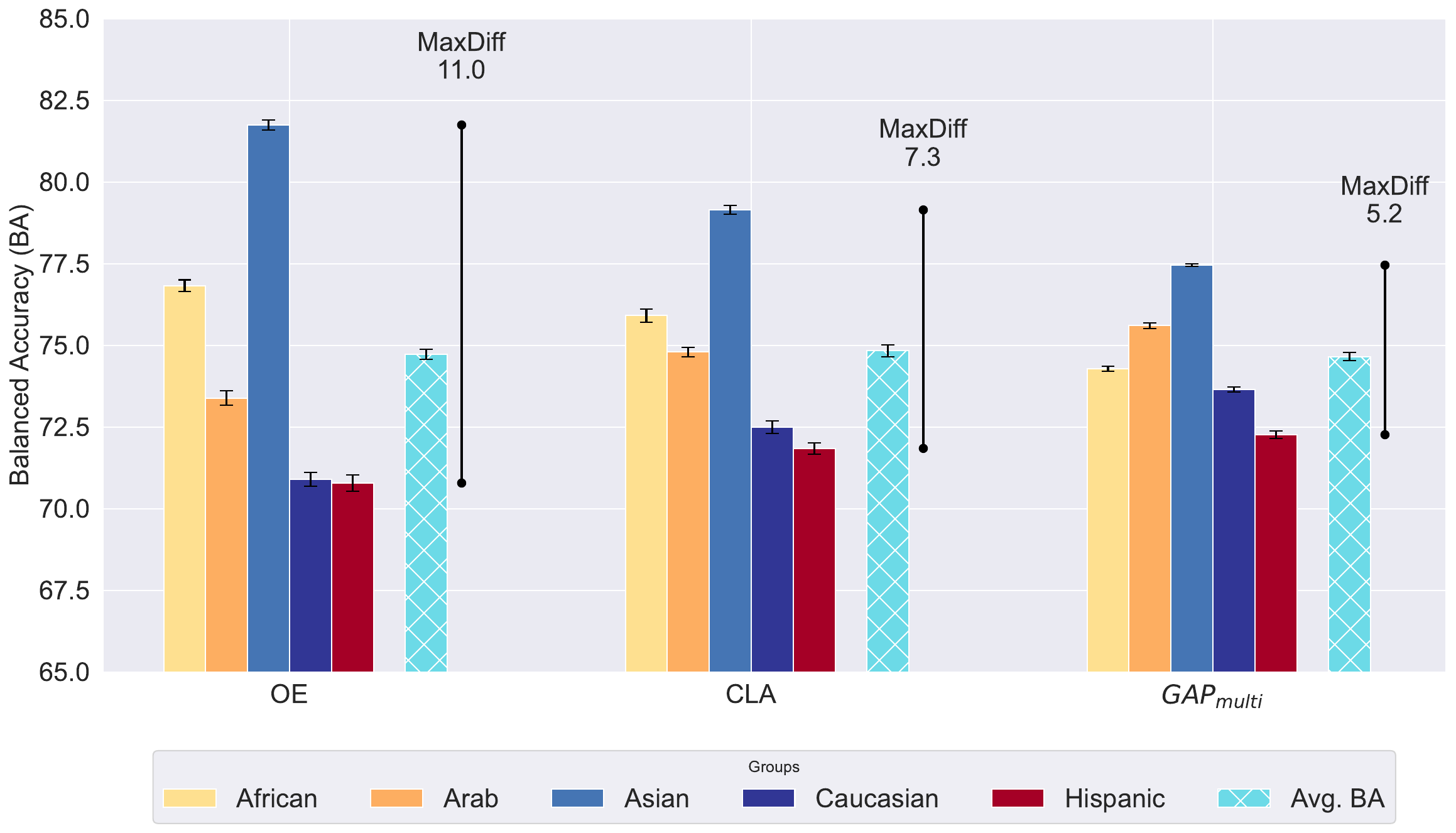}
    \vspace{-0.5em}
    \caption{Visualization of the BA values achieved by each loss over the 5 demographic groups in HateXplain, using DistilBERT as the text encoder. The maximum difference (Max. Diff.) of BA achieved for each loss across groups is also shown. 
    $GAP_{multi}$ performs best with lowest Max Diff. of 5.2, showcasing that it greatly reduces the disparities in performance across groups \vs other losses.}
    \label{fig:ba7group-hatexplain}
    \vspace{-0.75em}
\end{figure}

Fig. \ref{fig:ba7group-hatexplain} shows the visualization corresponding to Table \ref{tab:ba7group-hatexplain} in the main material on the 5-group HateXplain. For this dataset, CLA achieves the highest Avg. BA (74.84), but both OE and $GAP_{multi}$ achieve statistically comparable performance. Meanwhile, $GAP_{multi}$ achieves a  statistically significant reduction in Max. Diff.. The comparative reduction in disparity across groups, alongside similar Avg. BA, shows the method's effectiveness as a fair and utility-preserving loss function for multi target-group identification task.

\begin{figure}[h]
    \centering
    \includegraphics[width=\linewidth]{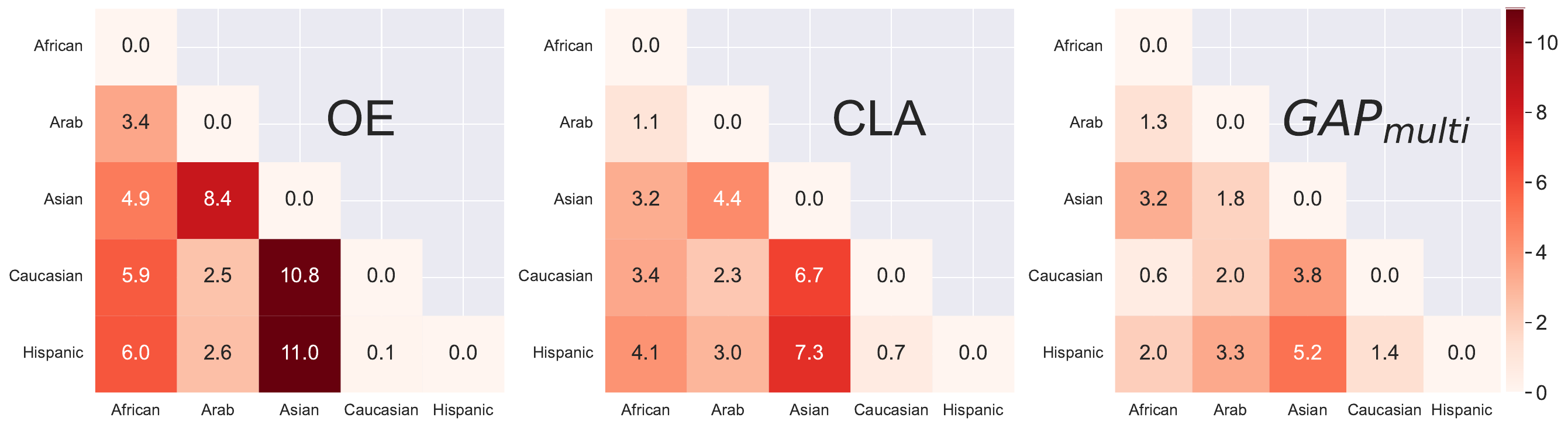}
    \vspace{-1.75em}
    \caption{Heatmap of pairwise absolute diff. of BA across groups in test set (HateXplain) as an indicator for bias and disparate impact. OE has the highest performance gap (Max Diff = 11.0) across groups indicated by the extremes of color, not only across one group-pair but across multiple group pairs. $GAP_{multi}$ has the least spread in pairwise error values (\textbf{Max. Diff. = 5.2}), evident from the flatness of color, indicating least disparate impact across groups.}
    \label{fig:pairwise-diff7group-hatexplain}
    \vspace{-0.75em}
\end{figure}

\textbf{Figure~\ref{fig:pairwise-diff7group-hatexplain}} displays pairwise absolute differences of Balanced Accuracy (BA) across various groups ($|\textit{left - bottom}|$) in HateXplain. Higher heatmap values indicate classifier bias, revealing consistent inequities, particularly for overall error (OE) rates, where the (\textit{African, Asian}) pair has the largest Max. Diff. (\textbf{11.0}). While CLA (optimized for FNR) improves over OE, its misaligned objective (minimizing false negatives, rather than equitable performance), results in a heatmap in-between that of OE and $GAP_{multi}$. In contrast, the $GAP_{multi}$ loss, explicitly designed for balanced group performance, shows substantially fewer extremes (\textbf{Max. Diff.: 5.2}), with smoother heatmap transitions indicating equitable classifier outcomes, aligning with our fair target-group detection objective.

\begin{figure}
    \centering
    \includegraphics[width=0.6\linewidth]{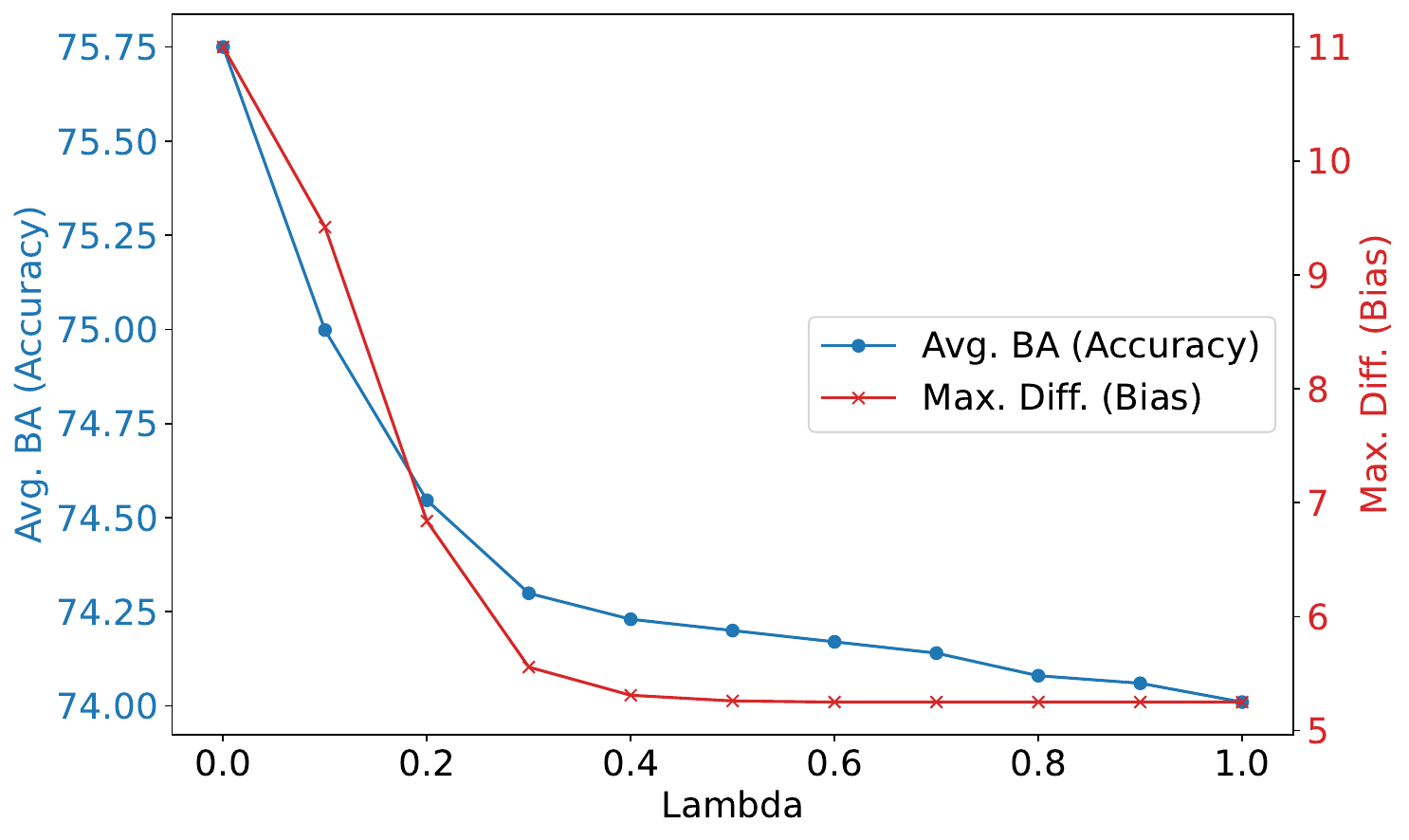}
    \vspace{-1.em}
    \caption{Hyperparameter Tuning Results showing variation of Avg. BA and Max. Diff. \vs $\lambda$ on HateXplain. For this dataset, with increasing values of $\lambda$ we observe a steady decrease in accuracy at the cost of bias.}
    \label{fig:tradeoff-hatexplain}
\end{figure}

Fig. \ref{fig:tradeoff-hatexplain} shows sensitivity of accuracy-fairness (Average Balanced Accuracy - Maximum Difference) \vs $\lambda$. The stopping criteria for this experiment was triggered at $\lambda$=0.4, the bias criteria stabilized at 5.25, with accuracy 74.23\%. Beyond that point accuracy keeps on slightly decreasing while the fairness criteria plateaus. Hence we stop at $\lambda$=0.4 since all the accuracy values to the right are dominated values, hence ignored. In contrast to what we observe for the MHS Corpus, this curve follows the standard optimization trend where decreasing bias comes at the expense of overall accuracy.

\section{Pairwise Group Classifiers on MHS} \label{app:2group}
 
We present the performance across four losses (OE, CLA, ADV and $GAP_{multi}$) for some of the 2-group case to show different scenarios and performances. Best values are bolded (higher for BA, lower for Diff.). Since ADV cannot accommodate multiple groups, we were unable to report ADV numbers for the 7-group case in the main material. ADV \cite{xia2020demoting} is an approximate adversarial loss for balancing FPR rates across groups. We notice similar issues of convergence instability as they observed in \citet{xia2020demoting} as well. Consequently, let ADV run for a fixed epochs and report the best BA value achieved over iterations.

\begin{table}[h]
\caption{Different 2-group setting highlighting the performance variation across the measures for different groups. (a) Optimizing across groups results in improving overall error, indicating that the group label provides an extra dimension for the loss to stabilize at a better local optima. (b) Performance varies across groups due to their population size, where the statistically major group dominates. (c) Groups have similar population, but performance varies due to one group being more difficult to model.}
\begin{minipage}{.33\textwidth}
    \centering
    \resizebox{\textwidth}{!}{
    \begin{tabular}{l|cc|cc} \toprule
          & \multicolumn{2}{c|}{Balanced Accuracy (BA)} & & \\ \midrule
         Loss & Latinx & Middle Eastern & Avg. BA & Diff. \\ \midrule
         OE & \textcolor{blue}{90.98} & \textcolor{red}{83.67} & 87.33 & 7.31 \\ 
         CLA & 91.52 & 88.73 & 90.12 & 2.79  \\
         ADV & 91.04 & 84.20 & 87.62 & 6.84 \\  \midrule
         $GAP_{multi}$ & \textcolor{blue}{\textbf{92.34}} & \textcolor{red}{\textbf{91.79}} & \textbf{92.06} & \textbf{0.55} \\ \bottomrule
    \end{tabular}
    }
    \subcaption{Case I}
    \label{tab:case1}
\end{minipage}
\begin{minipage}{.33\textwidth}
    \centering
    \resizebox{\textwidth}{!}{
    \begin{tabular}{l|cc|cc} \toprule
          & \multicolumn{2}{c|}{Balanced Accuracy (BA)} & & \\ \midrule
         Loss & Asian & Black & Avg. BA & Diff.\\ \midrule
         OE & \textcolor{red}{86.59} & \textcolor{blue}{\textbf{92.32}} & 89.46 & 5.73 \\ 
         CLA & 87.02 & 92.22 & 89.62 & 5.20 \\
         ADV & 87.49 & 92.10 & 89.79 & 4.61 \\ \midrule
         $GAP_{multi}$ & \textcolor{red}{\textbf{89.65}} & \textcolor{blue}{90.82} & \textbf{90.23} & \textbf{1.17} \\ \bottomrule
    \end{tabular}
    }
    \subcaption{Case II}
    \label{tab:case2}
\end{minipage}
\begin{minipage}{.33\textwidth}
    \centering
    \resizebox{\textwidth}{!}{
    \begin{tabular}{l|cc|cc} \toprule
          & \multicolumn{2}{c|}{Balanced Accuracy (BA)} & & \\ \midrule
         Loss & White & Latinx & Avg. BA & Diff. \\ \midrule
         OE & \textcolor{blue}{\textbf{88.89}} & \textcolor{red}{82.19} & 85.54 & 6.70 \\ 
         CLA & 88.20 & 85.55 & 86.87 & 2.65 \\
         ADV & 88.18 & 82.19 & 85.18 & 5.99 \\ \midrule
         $GAP_{multi}$ & \textcolor{blue}{88.36} & \textcolor{red}{\textbf{86.23}} & \textbf{87.29} & \textbf{2.13} \\ \bottomrule
    \end{tabular}
    }
    \subcaption{Case III}
    \label{tab:case3}
\end{minipage}
\end{table}

\end{document}